\documentclass[3p,times]{elsarticle}

\usepackage{textcomp}
\usepackage{url}
\usepackage{natbib}
\usepackage{hyperref}
\usepackage{threeparttable}
\usepackage{xcolor}

\usepackage{array}
\usepackage{multirow}
\usepackage{multicol}
\usepackage{makecell}
\usepackage{pgfplots}
\pgfplotsset{compat=1.5}
\usetikzlibrary[patterns]
\usepackage{subcaption}
\usepackage{graphicx}

\def\debug{}

\ifdefined\debug

\else

\fi

\pgfdeclarepatternformonly{south west lines}{\pgfqpoint{-0pt}{-0pt}}{\pgfqpoint{3pt}{3pt}}{\pgfqpoint{3pt}{3pt}}{
        \pgfsetlinewidth{0.4pt}
        \pgfpathmoveto{\pgfqpoint{0pt}{0pt}}
        \pgfpathlineto{\pgfqpoint{3pt}{3pt}}
        \pgfpathmoveto{\pgfqpoint{2.8pt}{-.2pt}}
        \pgfpathlineto{\pgfqpoint{3.2pt}{.2pt}}
        \pgfpathmoveto{\pgfqpoint{-.2pt}{2.8pt}}
        \pgfpathlineto{\pgfqpoint{.2pt}{3.2pt}}
        \pgfusepath{stroke}}

\begin{document}

%%
%% The "title" command has an optional parameter,
%% allowing the author to define a "short title" to be used in page headers.
\title{Fine-grained TLS services classification with reject option}

%%
%% The "author" command and its associated commands are used to define
%% the authors and their affiliations.
%% Of note is the shared affiliation of the first two authors, and the
%% "authornote" and "authornotemark" commands
%% used to denote shared contribution to the research.

\author[cesnet,fit]{Jan Luxemburk}
\ead{luxemburk@cesnet.cz}

\author[cesnet]{Tomáš Čejka}
\ead{cejkat@cesnet.cz}

\address[cesnet]{CESNET, Czech Republic}
\address[fit]{Faculty of Information Technology, CTU in Prague, Czech Republic}

\journal{Computer Networks}
\let\today\relax

%%
%% The abstract is a short summary of the work to be presented in the
%% article.
\begin{abstract}
The recent success and proliferation of machine learning and deep learning have provided powerful tools, which are also utilized for encrypted traffic analysis, classification, and threat detection in computer networks. These methods, neural networks in particular, are often complex and require a huge corpus of training data. Therefore, this paper focuses on collecting a large up-to-date dataset with almost 200 fine-grained service labels and 140 million network flows extended with packet-level metadata. The number of flows is three orders of magnitude higher than in other existing public labeled datasets of encrypted traffic. The number of service labels, which is important to make the problem hard and realistic, is four times higher than in the public dataset with the most class labels. The published dataset is intended as a benchmark for identifying services in encrypted traffic. Service identification can be further extended with the task of ``rejecting'' unknown services, i.e., the traffic not seen during the training phase. Neural networks offer superior performance for tackling this more challenging problem. To showcase the dataset's usefulness, we implemented a neural network with a multi-modal architecture, which is the state-of-the-art approach, and achieved 97.04\% classification accuracy and detected 91.94\% of unknown services with 5\% false positive rate.
\end{abstract}


%%
%% This command processes the author and affiliation and title
%% information and builds the first part of the formatted document.
\maketitle
 
\section{Introduction}
In this work, we target the topic of encrypted traffic classification and elaborate on the task of identification of web services running in HTTPS communications. Network operators leverage information about web services for firewall decisions, QoS prioritization, and enforcement of network-usage policies. Because most of the network traffic is being encrypted, the traditional deep-packet-inspecting (DPI) solutions are becoming obsolete, and there is an urgent need for modern classification methods capable of analyzing encrypted traffic. These methods have to forgo the packet's opaque payload and focus on other types of information about network traffic. There are several available traffic characteristics, such as flow statistics and sequences of packet sizes, directions, and inter-arrival times. Packet sizes, in particular, were shown to be a rich source of information for traffic classification, and neural networks are the state-of-the-art approach for processing them~\cite{aceto_MIMETICMobile_2019, rezaei_LargescaleMobile_2020, yang_DeepLearning_2021, lopez-martin_NetworkTraffic_2017}. However, neural networks require a huge corpus of training data.

Therefore, we have created a dataset spanning two weeks, consisting of 140 million network flows, and having 200 web services. We use TLS Server Name Indication (SNI), which is a domain name present in cleartext in TLS communications, to assign ``ground-truth'' labels of the 200 services. For that, we created an SNI-service mapping, where each service is mapped to one or multiple SNI domains. The selection of services and creation of the SNI-service mapping are described in Section \ref{sec:service-definition}. The dataset collection environment is described in Section \ref{dataset-collection}. A comparison to existing datasets is presented in Section~\ref{sec:relatedwork}, which also describes related works. 

Using the created dataset, we train a neural network classifier based on 1D convolutional and linear layers. The model is described in Section \ref{sec:model}. Following the recent literature \cite{yang_DeepLearning_2021, rezaei_DeepLearning_2019}, we implemented a reject option that gives the model an option to abstain from classification for ambiguous, uncertain, or out-of-distribution samples. This technique is leveraged to identify unknown services---those that were not present in the training dataset. We consider the reject option to be essential for the practical deployment of traffic classification models because network traffic training datasets cannot be complete in the sense of containing all possible classes (web services in our case). Internet is a dynamic environment with new apps and services introduced each week, and it does not make much sense to force the classifier to make decisions on these novel classes while, in fact, there are techniques that enable discovering them. Detection of unknown services can be especially important in restricted networks, where unknown traffic means potential intrusion or violation of policies. The reject option and methods for novel class detection are described in detail in Section~\ref{sec:reject-option}. The experimental setup is described in Section~\ref{sec:experimental-setup}, and the evaluation workflow and the used performance metrics in Section~\ref{sec:evaluation}. A description of achieved results, a feature importance evaluation, and model interpretations are provided in Section~\ref{sec:results}.

\medskip\noindent
The main contributions are:
\begin{itemize}
    \item We create and publish a large 140-million-flow dataset of TLS traffic, having the highest number of classes among available public datasets intended for encrypted network traffic classification.
    \item To showcase the dataset's usefulness, we train a multi-modal neural network based on 1D convolutions and linear layers capable of identifying web services in encrypted traffic. The model's input data are packet metadata sequences and flow statistics, which are the standard features used for encrypted traffic analysis. We also provide an analysis of the influence of individual input features and explainable AI (XAI) interpretations of the trained neural network.
    \item We implement and compare three novel class detection (reject) methods that enable discovering unknown services that were not present in the training dataset. An energy-based approach, which offers a balance between detection rate and inference speed, is used for the first time in the traffic classification domain.
\end{itemize}

\section{Dataset}
\begin{figure}[ht]
	\centering
	\includegraphics[width=0.75\linewidth]{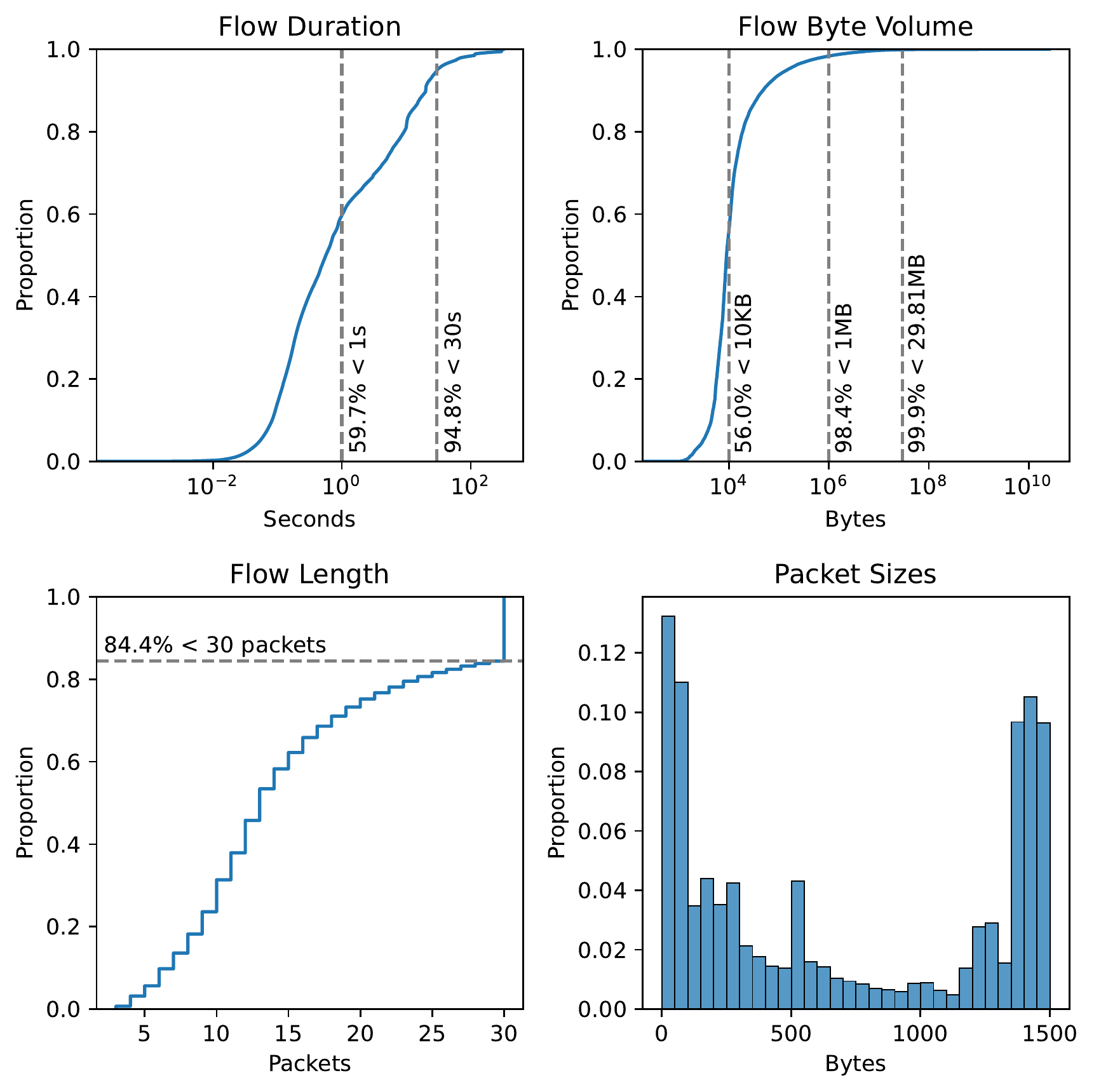}
	\caption{Dataset statistics overview showing cumulative distribution functions of flow duration, byte volume, and flow packet length. A histogram (50 bin size) of packet sizes across the whole dataset is also shown. Almost 60\% of flows are shorter than one second; 56\% of flows have less than 10 kilobytes; and the median flow length is 13 packets. 84.4\% of flows are shorter than 30 packets and are therefore padded in preprocessing (explained more in Section~\ref{sec:model-architecture}). Packet sizes tend to be either in the range close to the 1500 bytes MTU, which is the consequence of splitting larger application-layer messages into MTU-sized chunks, or small up to 100 bytes, which can be attributed to small HTTPS requests, application keep-alives, etc. We attribute the small peak in 500-550 bytes to packets carrying the TLS ClientHello messages.}
	\label{fig:dataset-stats}
\end{figure}

Our dataset is built from live traffic observed within the first two weeks of October 2021 using high-speed monitoring probes at the perimeter of the CESNET2 network. CESNET2 is a national research and education network of the Czech Republic that has 500 connected institutions with about half a million users in total. The dataset contains 140 million flow records in total that were exported from multiple 100\,Gb/s network links. This capture environment provides realistic characteristics of traffic originating from various web browsers, operating systems, mobile devices, desktop machines, and both HTTP/1.1 and HTTP/2 protocols. Also, due to the organic nature of the captured traffic, the dataset includes rich behaviors of each web service, such as different user actions and settings within each service. Dataset statistics about the distributions of flow duration, byte volume, and length are shown in Figure~\ref{fig:dataset-stats}.

We decided to make the dataset public\footnote{The dataset is available for download at \url{https://www.liberouter.org/datasets/cesnet-tls22}.} to address the need for open datasets with a large number of traffic classes, which was presented in the literature \cite{aceto_MIRAGEMobileapp_2019,shahraki_ActiveLearning_2021,yang_DeepLearning_2021}. We also created a guideline on how to replicate the dataset collection process in other environments. The guideline, together with a Dockerfile and example configuration files, is available at the dataset download page.

\subsection{Dataset collection process}
\label{dataset-collection}
The dataset collection was performed using open-source tools running within the production monitoring infrastructure. Since the creation of the dataset is based on extended IP flow records, the first tool was an IPFIX flow exporter called ipfixprobe\footnote{\url{https://github.com/CESNET/ipfixprobe}}. The traditional IP flow data (communicating IP addresses, transport ports and protocol, timestamps, the sum of bytes and packets) were further extended with ipfixprobe plugins providing advanced field parsing and export. The PSTATS plugin provides the most important exported features---the metadata from the first 30 packets of each bi-directional connection. The annotation of the dataset was based on TLS information from the TLS plugin, which parses SNI domains from TLS handshakes.

Flow data from multiple instances of monitoring probes were collected at a central flow collector IPFIXcol2\footnote{\url{https://github.com/CESNET/IPFIXcol2}}, converted into an efficient binary data format, and passed into the NEMEA\footnote{\url{https://github.com/CESNET/NEMEA}} system for further processing \cite{cejka_NEMEAframework_2016}. For the purpose of collecting this dataset, we have developed a NEMEA module called SNI-dataset-saver to perform filtering, matching, and sampling of flow data. SNI-dataset-saver uses a trie data structure for efficient lookup of domain names and ``wildcards'' (i.e., partial domain names like ``*.youtube.com''). After matching a domain from a service's list of domains/wildcards, the module performs appropriate sampling and stores the data into compressed files representing time windows (we chose the traditional 5-minute interval to rotate the files). The overall process of dataset collection is visualized in Figure \ref{fig:data-processing}. The rest of this section describes individual steps of the dataset collection process.

\subsubsection{TLS handshake flows}
A TLS connection starts with a TLS handshake for server authentication and the establishment of an encrypted tunnel; after the handshake, application data is exchanged. Our dataset consists only of TLS flows that contain the handshake; other flows are filtered. Due to practical reasons, flow exporters split long communications into multiple flow records. A typical setting is to limit the maximum duration of flow to 300 seconds; after this time, the flow is exported (and a new flow record is created if the communication continues). Classification of TLS handshake flows is suitable for service identification because the label predicted for the handshake flow can be reused for the rest of the TLS communication (i.e., if the communication is longer than 300 seconds and multiple flow records are created for it, the label predicted for the first handshake flow can be reused for the subsequent flows).

\subsubsection{Flow sampling}
The data collection process performs dynamic flow sampling; flows of services with the most samples are sampled in a ratio of 1:15, flows of the least frequent services are not sampled at all, flows of services in between are sampled in a ratio ranging from 1:2 to 1:9. The order of services is updated during dataset collection; thus, the sampling ratio for a given service can change in time. The reasoning is to soften the huge imbalances between the top 20 services and the rest. The top couple of services, such as ads networks traffic or Windows updates, account for a disproportionate number of flows. For the purpose of training the 200-services classifier, it is suitable to include all samples of the less prevalent services and omit some samples of the top services. The second reason for the sampling is practical; it is not feasible to store all the data because of infrastructure and storage size limitations and the enormous volume of monitored traffic in the ISP backbone network.

\subsubsection{Removing flows with less than three packets}
We omit TLS flows with less than three packets, which is the minimum number of packets that a valid TLS connection should have according to the RFC specifications. The shortest three-packet TLS connection is possible with a 0-RTT handshake in TLS 1.3. Older versions of TLS require more packets to be exchanged during the handshake.

\subsubsection{Removing uni-directional flows}
\label{asymmetric-routing}
We omit flows with missing one direction of the communication. These flows occur when communications use different network paths for uplink and downlink directions, and one of the paths is not monitored. This can happen when an organization connected into CESNET2 has multiple peerings, and due to asymmetric routing, one direction of the communication is transferred via a different ISP.

\subsubsection{Data anonymization}
We stripped the dataset of communicating IP addresses, transport ports, absolute timestamps, and raw TLS SNI domains; of course, the ground-truth labels derived from the TLS SNI are included. With those user-identifying fields removed, there is no context information attached to samples, and user de-anonymization attacks cannot be performed on the dataset traffic. For privacy and ethical reasons, we have never stored the payload of monitored communications during the dataset collection.

\subsection{Service definition}
\label{sec:service-definition}
For building our dataset, we selected 200 web services. For each service, we collected a list of domains it is using. The criteria for selecting services were: 

\begin{enumerate}
    \item The main goal was to cover diverse services. We have defined 21 service categories, almost each with at least five representatives. The distribution of services, flows, and bytes volume across the service categories is shown in Figure \ref{fig:dataset-categories}. The top 100 service labels can be found in \ref{appendix:classification-report}.
    \item We did not include static websites with almost no user interactions, such as wikis or news sites. Identification of those websites is more suitable for website fingerprinting approaches.
    \item We prioritize services with more traffic volume so that the model trained on the dataset covers a large proportion of the monitored traffic.
\end{enumerate}

To find the domains associated with a service, we searched its online documentation. In some cases, we found a docpage about ``whitelist domains for network and firewall settings'', which contained all the service's domains. In some cases, we used Netify's Application Lookup Tool\footnote{\url{https://www.netify.ai/resources/applications}}. For the rest, we analyzed the monitored SNI values and handpicked the domains. The created SNI-service mapping is available at the dataset download page.

We also organize services with the same provider into groups, such as Google's services or Mozilla's services, and later use this hierarchy to compute ``superclass'' performance metrics. This is valuable in scenarios in which we do not mind misclassifying services among the same provider group; in other words, a predicted label ``a Google service'' is good enough for the network operator. Note that we have included both fine-grained services, such as ESET Push Notifications or Opera Autoupdate, and coarse-grained, such as Instagram, Outlook, or Netflix. Fine-grained services can be seen as single-function services/APIs; on the other hand, coarse-grained services can be composed of many different sub-functionalities.

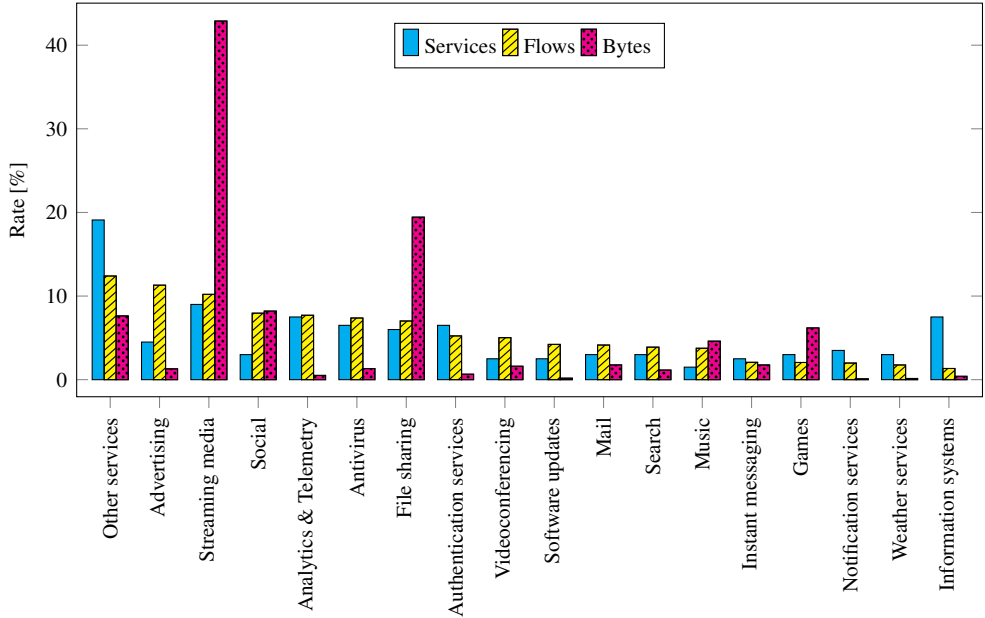
\begin{figure}[t]
\centering
%%%%
\begin{tikzpicture}
\begin{axis}[
    tick label style={font=\footnotesize},
    label style={font=\footnotesize},
    width=13.5cm,
    height=6.8cm,
    enlarge y limits=0.05,
    enlarge x limits=0.04,
    ylabel={Rate [\%]},
    xlabel={},
    symbolic x coords={
        Other services,
        Advertising,
        Streaming media,
        Social,
        Analytics \& Telemetry,
        Antivirus,
        File sharing,
        Authentication services,
        Videoconferencing,
        Software updates,
        Mail,
        Search,
        Music,
        Instant messaging,
        Games,
        Notification services,
        Weather services,
        Information systems,
        Internet banking,
        Remote desktop,
        Virtual assistant},
    xtick=data,
    legend style={at={(0.5,0.95)},
     anchor=north,legend columns=-1,
     font=\footnotesize},
    ybar=0pt,
    bar width=4.5pt,
    xticklabel style={rotate=90, anchor=east},
    xtick pos=bottom,
    legend image code/.code={
        \draw [#1] (0cm,-0.1cm) rectangle (0.2cm,0.25cm); },
    ]
    % \tikzset{every node}=[font=\sffamily]
    \addplot [fill=cyan] coordinates {
    (Other services, 19.1) (Advertising, 4.5) (Streaming media, 9) (Social, 3) (Analytics \& Telemetry, 7.5) (Antivirus, 6.5) (File sharing, 6) (Authentication services, 6.5) (Videoconferencing, 2.5) (Software updates, 2.5) (Mail, 3) (Search, 3) (Music, 1.5) (Instant messaging, 2.5) (Games, 3) (Notification services, 3.5) (Weather services, 3) (Information systems, 7.5) %(Internet banking, 3.5) (Remote desktop, 1.5) (Virtual assistant, 1)
    };
    \addplot [black,fill=yellow, postaction={pattern=south west lines}] coordinates {
    (Other services, 12.398) (Advertising, 11.301) (Streaming media, 10.205) (Social, 7.954) (Analytics \& Telemetry, 7.704) (Antivirus, 7.376) (File sharing, 7.018) (Authentication services, 5.234) (Videoconferencing, 5.023) (Software updates, 4.221) (Mail, 4.143) (Search, 3.89) (Music, 3.758) (Instant messaging, 2.076) (Games, 2.058) (Notification services, 1.991) (Weather services, 1.759) (Information systems, 1.34) %(Internet banking, 0.355) (Remote desktop, 0.173) (Virtual assistant, 0.026)
    };
    \addplot [black, fill=magenta, postaction={pattern=crosshatch dots}] coordinates {
    (Other services, 7.626) (Advertising, 1.297) (Streaming media, 42.888) (Social, 8.204) (Analytics \& Telemetry, 0.513) (Antivirus, 1.331) (File sharing, 19.436) (Authentication services, 0.656) (Videoconferencing, 1.61) (Software updates, 0.181) (Mail, 1.766) (Search, 1.152) (Music, 4.607) (Instant messaging, 1.763) (Games, 6.189) (Notification services, 0.101) (Weather services, 0.144) (Information systems, 0.405) %(Internet banking, 0.126) (Remote desktop, 0.005) (Virtual assistant, 0.002)
    };
    \legend{Services,Flows,Bytes}
\end{axis}
\end{tikzpicture}
%%%
	\caption{A breakdown of dataset traffic into service categories, showing fractions of services, bytes and flows. We omit three categories with the smallest contributions: \textit{Internet banking} (3.5\% services, 0.4\% flows, 0.15\% bytes),\textit{ Remote desktop} (1.5\% services, 0.2\% flows, 0.01\% bytes), and \textit{Virtual assistant} (1\% services, 0.03\% flows, 0.01\% bytes).}
	\label{fig:dataset-categories}
\end{figure}

Out of the 200 services, 59 are Czech services (due to the dataset collection location), others are international. With those 200 selected services, the dataset represents 45\% of all TLS flows with SNI observed in the production network on port 443/TCP. We consider this coverage sufficient for research purposes.

\subsection{Discussion on usage of TLS SNI}
\label{sec:discussion-sni}
We present three viewpoints on how a generalizing model trained on our dataset compares to using a deterministic SNI-service map (like the one used to label the dataset).

\begin{itemize}
    \item As noted in \cite{shbair_ImprovingSNIBased_2016}, SNI can be faked to avoid SNI-based filtering. 
    This renders the deterministic mapping useless, for example, for the detection of malware faking SNI to hide its TLS communication.
    \item SNI might be encrypted in the near future. There is an RFC defining TLS extension called Encrypted Client Hello, which encrypts the entire Client Hello message. It can be seen as a continuation of the TLS 1.3 effort that moved most of the cleartext metadata (e.g., certificates) of the TLS handshake into the encrypted tunnel. If this extension got widespread adoption, monitoring of TLS traffic would be limited to a bare minimum because TLS handshakes would become opaque to network operators. In this scenario, a generalizing model, like the one proposed in Section \ref{sec:model}, could replace the lost SNI information. Our labeling procedure would be usable as long as a portion of monitored TLS traffic does NOT encrypt its SNI; when all SNIs are encrypted, the last resort option would be to collect the training dataset on TLS proxies or endpoint devices.
    \item SNI-service mapping can 
    never be complete to cover all domains and services on the Internet and is cumbersome to keep up-to-date. A model based on the behavior of packet sequences can work as a generalization of the mapping and could also discover services on unknown domains. An example scenario: a model is trained on several domains hosting Microsoft Outlook; however, the list of Outlook domains is not complete, and thus using the list for service identification would miss some Outlook communication. On the other hand, the model can learn the distinctive behavior of Outlook communication and discover it regardless of domains. 
\end{itemize} 

Moreover, both the deterministic SNI-service mapping and the behavior generalizing model can work in cooperation. The model can be used as a safe-check for the SNI; an anomaly indicator can be raised whenever the behavior of monitored communication and the model's prediction does not match the service deduced from the observed SNI. 

\section{TLS services classification with reject option}
We believe the presented dataset can be helpful for traffic classification research communities. Among the tasks that can be benchmarked and improved upon on this dataset are:

\begin{enumerate}
\item Service identification in encrypted traffic with or without detection of unknown traffic.
\item Traffic type classification, with categories such as videoconferencing, web browsing, or software updates.
\item Classification whether a network flow is a result of human interaction or of some automation, such as periodic API calls, software updates, or adds distribution.
\end{enumerate}

In the following sections, we elaborate on task (1), with the main focus on the detection of unknown traffic. We focus on this harder problem because of its relevance for the practical deployment of ML-based classifiers of network traffic. The target task is the identification of web services in encrypted TLS traffic with the detection of unknown services. To summarize, the goal of the model is to process traffic characteristics of encrypted TLS communications in order to predict the web service---or, if the communication is not similar to any service from the training set, label the communication as \textit{unknown}. 

We describe our solution in the  three sections: the input data in Section~\ref{sec:input}, the chosen model in Section \ref{sec:model}, and the methods for detection of unknown services in Section~\ref{sec:reject-option}.

\subsection{Input}
\label{sec:input}
Input data are of two types: packet sequences (PSTATS) and flow statistics (FLOWSTATS). We share a similar attitude as in \cite{yang_DeepLearning_2021} and opt for data types, which are reusable across different networks and for other use-cases and classification tasks (which is desirable from a scientific standpoint), and avoid using host-related information, such as autonomous system numbers of the destination hosts (which would be desirable for the practical deployment).

\subsubsection{Packet sequences}
Sequences of packet sizes, directions, and inter-arrival times (IAT) are standard data input for traffic analysis. For the packet sizes, we consider payload size after TCP headers. We omit packets with no TCP payload, for example ACKs, because zero-payload packets are related to the transport layer internals rather than services' behavior. Packet directions are encoded as \textpm1. Packet inter-arrival times depend on the location of communicating hosts, their distance, and on the network conditions on the path. However, the experiments prove that it is possible to extract relevant information that correlates with user interactions and, for example, with the time required for an API/server/database to process the received data and generate the response to be sent in the next packet. As noted in Section \ref{dataset-collection}, PSTATS include information about the first 30 packets of each flow. The model never sees the actual bytes of packets' payload.

\subsubsection{Flow statistics}
The other type of input data, flow statistics, contain aggregated information about the entire bi-directional flow. We use the standard fields: the number of transmitted bytes and packets in both directions, the duration of flow, and the number of roundtrips. We count roundtrips as the number of changes in the communication direction (from packet directions data); in other words, each client request and server response pair is one roundtrip. We also use the presence of individual TCP flags in both directions, however, we omit flags that have no variance and are either present in all flows (SYN, ACK) or almost in none (URG). The list of used flow statistics can be seen in Figure~\ref{fig:xai-flowstats}.

\subsubsection{Data preprocessing}
It is recommended to standardize input features before training a neural network. We standardize packet sizes and times using mean and variance (z-score). For flow statistics, which do not have a fixed range (the flow duration, the sum of bytes, etc.), we opt for robust scaling that uses median and inter-quartile range instead of mean and variance. This limits the negative influence of outliers. Moreover, before standardization, we also clip flow statistics to their 0.95 quantiles to further reduce the influence of extreme values. Packet times are clipped between 1\,millisecond and 15\,seconds. Packet sizes have a range of 1 (zero-payload packets are omitted) to 1460 bytes (the typical TCP maximum segment size). The TCP flags are 0/1 features and are not standardized. Directions of packets are encoded as \textpm1 and are not standardized.

\begin{figure}[ht]
	\centering
	\includegraphics[width=0.67\linewidth]{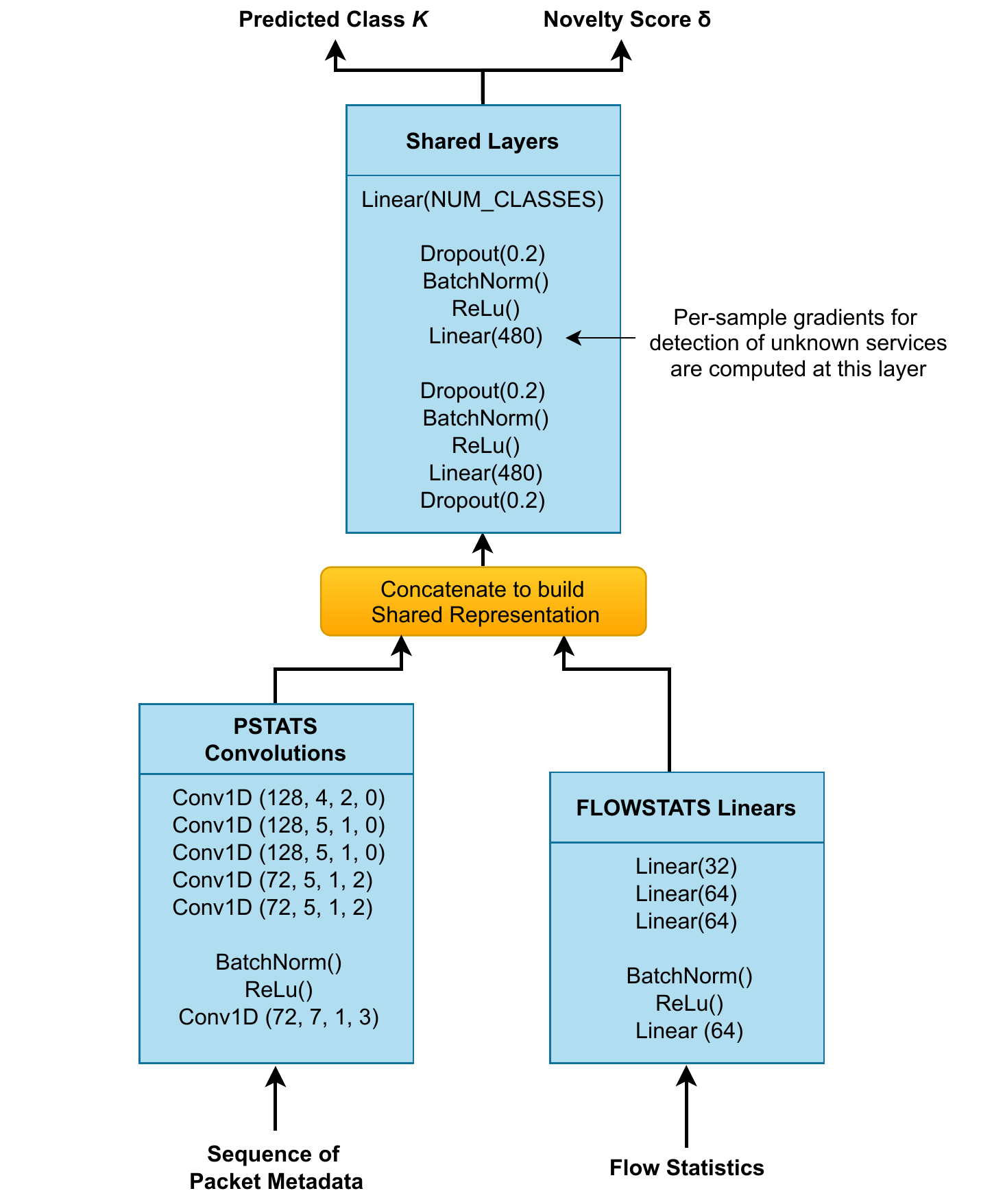}
	\caption{The neural network multi-modal architecture. The layer parameters
represent: \texttt{Conv1D(\#filters, kernel\_size, stride, padding), Linear
(\#out\_features), Dropout(rate)}. Some \texttt{BatchNorm} and \texttt{ReLu} layers are omitted to save space.}
	\label{fig:model-architecture}
\end{figure}

\subsection{Model}
\label{sec:model}
According to the experiments of Yang~et~al.~(\cite{yang_DeepLearning_2021}, the paper is described in detail in Section~\ref{sec:relatedwork}), standard ML and DL offer comparable performance in encrypted traffic classification tasks. The authors have even shown that the XGBoost model, which is a tree-based boosting ensemble model, can be tuned to beat the classification performance of a neural network. We have performed a similar experiment and compared LightGBM, which is also a boosting model based on decision trees, with the neural network presented in the next section. The outcome of this experiment is that our neural network provides better classification performance than the ML model. Discussion and detailed results of this comparison are presented in Section~\ref{sec:ML-DL-comparison}. 

ML and DL models, however, differ in their abilities to detect novel classes (unknown traffic), because the non-linear feature extraction process of neural networks backbone provides sizeable advantages for this task~\cite{yang_DeepLearning_2021}. Moreover, DL-based models are better suited for transfer learning techniques, which are useful in scenarios when a trained model is reused for another similar task for which the training data is scarce (for example, encrypted malware traffic detection).

% Because of our focus on novel class detection, the better classification performance that neural networks offer, and based on the conclusions of related work, we chose a deep neural network based on 1D convolutions as our model.

Based on the provided rationale, the conclusions of related work, the better classification performance, and because of our focus on novel class detection, we chose a deep neural network based on 1D convolutions as our model.

\subsubsection{Model architecture}
\label{sec:model-architecture}
Our neural network has two separate chains of layers for processing packet sequences and flow statistics; outputs of those are concatenated to create a shared representation and processed with more layers until the final ``classification'' layer and softmax function. The network is visualized in Figure \ref{fig:model-architecture}. This network architecture, which is also used in \cite{aceto_MIMETICMobile_2019,akbari_LookCurtain_2021}, is called multimodal and is studied in detail in \cite{aceto_DISTILLEREncrypted_2021}. We have two modalities, two types of input data, and each is processed with a different chain of layers. Packet sequences are processed with 1D convolutions; on the other hand, flow statistics are not ``sequence-like'' and are processed with simple linear layers. PSTATS features are padded with zeros to the maximum length of 30 because convolutions require a fixed length of the input. Batch normalization is used in both modalities, and dropout is employed as a regularization method in the shared representation part. ReLu is used as the activation function. We experimented with pooling layers without much success. The final model has 1.1 million trainable parameters.

\subsection{Reject option \& detection of unknown services}
\label{sec:reject-option}
In the context of traffic classification, the so-called ``reject option'' was employed in \cite{aceto_MobileEncrypted_2019} for increasing a model's prediction performance at the expense of not labeling some of the samples. In ML literature, this is sometimes called selective classification. The principle is that we allow the model to abstain from classification for ambiguous, uncertain, or out-of-distribution samples. This approach was justified in \cite{taylor_RobustSmartphone_2018} for the task of identifying mobile apps. Since apps typically create multiple flows when used, there remains a high chance of identifying them from their more distinctive flows, without the need to classify all the flows.

An almost identical technique, but with a different goal in mind, is employed in \cite{yang_DeepLearning_2021}. The challenge here is to discover samples of new classes that were not present in the training dataset. This problem is called \textbf{novel class detection} in ML literature and is the focus of this work. The core idea is to compute a \textit{novelty score} for each sample we want to classify; when the score is greater than a decision threshold, the predicted label is replaced with an \textit{unknown label}. The value of the decision threshold is computed over a validation set. The typical setting is to calculate the 95th percentile of validation samples` \textit{novelty scores}; in other words, the threshold is set to reject 5\% of known samples as unknown (5\% False Positive Rate). 

We implemented three methods for novel class (NC) detection: a baseline softmax score method \cite{hendrycks_BaselineDetecting_2018}, energy-based out-of-distribution detection \cite{liu_EnergybasedOutofdistribution_2021}, and a gradient-based rejection technique introduced in \cite{yang_DeepLearning_2021}.

\subsubsection{Baseline --- softmax score}
The naive baseline is taking the (negated) maximum of the softmax output as the \textit{novelty score}. Denoting with $x$ the logits (the output of the final layer of size $N$), the softmax value for class $k$ is defined in Equation \ref{eq:softmax}. Using the definition in Equation \ref{eq:softmax}, the softmax score is computed as $\max_k P(y = k | x)$. This approach, however, is suboptimal for NC detection because neural networks can produce overconfident predictions (high softmax probabilities) on inputs far away from training data \cite{nguyen_DeepNeural_2015}, which is the exact opposite of what we expect from a \textit{novelty score}. Softmax scores are negated to be consistent with the other methods in the sense that a higher \textit{novelty score} means more abnormal, novel input.

\begin{equation}
\label{eq:softmax}
P(y = k | x) = \frac{\exp{x_k}}{\sum_{i=i}^N \exp{x_i}}
\end{equation}

\subsubsection{Energy-based}
A simple yet effective improvement over the baseline is the energy-based method introduced in \cite{liu_EnergybasedOutofdistribution_2021}. It defines the \textit{novelty score} as free energy, which is calculated with the \texttt{LogSumExp} operator (also negated as the softmax score) over the logits vector denoted as $x$:

\begin{equation}
\label{eq:logsumexp}
\mathrm{LSE}(x_1, \dots, x_N) = \log \left(\sum_{i=1}^N\exp{x_i}\right) 
\end{equation}

\subsubsection{Gradient-based}
\label{sec:gradient-based}
The third implemented method is based on per-sample gradients. The idea is to pretend the prediction was correct and perform a backpropagation step with the predicted label as ground truth. After backpropagation, the p-norm of the gradient of the penultimate layer is used as the \textit{novelty score}. Apart from their interpretation as model updates, gradients can be seen as a measure of how familiar the model is with given inputs. The model is a converged solution, and the required model updates should be smaller for samples similar to training data than for samples not seen during training. Note that the weights of the neural network are never changed during inference. This method is detailed in works focused on traffic analysis \cite{yang_DeepLearning_2021,yang_ThinkbackTaskSpecific_2021}, but is also elaborated in general DL literature \cite{lee_GradientsMeasure_2020}. 

We extend this method; instead of using the standard cross-entropy loss to compute gradients, we use the SimLoss loss function \cite{kobs_SimLossClass_2020}. SimLoss is a modification of the cross-entropy loss designed for hierarchical classification. Cross-entropy assumes that one target class is correct; SimLoss changes this assumption and allows the definition of a matrix of similarities between classes. This matrix is used for loss computation such that mistakes between similar classes are less penalized. We use the organization of services into service groups to construct the matrix. The resulting effect for gradient-based NC detection is that samples with ambiguous predictions between services of the same group produce smaller loss, thus smaller gradients, and thus lower \textit{novelty scores}. As an example, a Facebook Web sample that is classified as 70\% Facebook Web, 25\% Facebook Messenger, and 5\% Facebook Media receives a lower \textit{novely score} than it would if the mistakes were non-Facebook services. As a result, there is a lower chance of misclassifying samples like this as \textit{unknown}.

The gradient-based method has three hyperparameters in total. The $p$ for computing p-norms of gradients (one is the Manhattan norm, two is the Euclidean norm), a coefficient $\alpha$ defining how much are services in the same group similar for SimLoss, and temperature parameter $T$ that is used for temperature scaling of softmax probabilities. Temperature scaling  \cite{guo_calibrationmodern_2017} is a de-facto standard method for calibration of neural network output. We found out that temperature scaling boosts the performance of softmax-score and gradient-based novel class detection methods. We do not use temperature scaling for the energy-based method because its authors do not recommend it.

The inference speed, which is crucial for monitoring of high-speed networks, is 5x slower for the gradient-based method than for the energy-based method, which is almost as fast as the softmax baseline. It is worth noting that no optimizations were elaborated for the measurement of the gradient method, which is more complex than the energy-based method and has more room for speed optimizations.

\begin{figure}[t]
	\centering
	\includegraphics[width=0.8\linewidth]{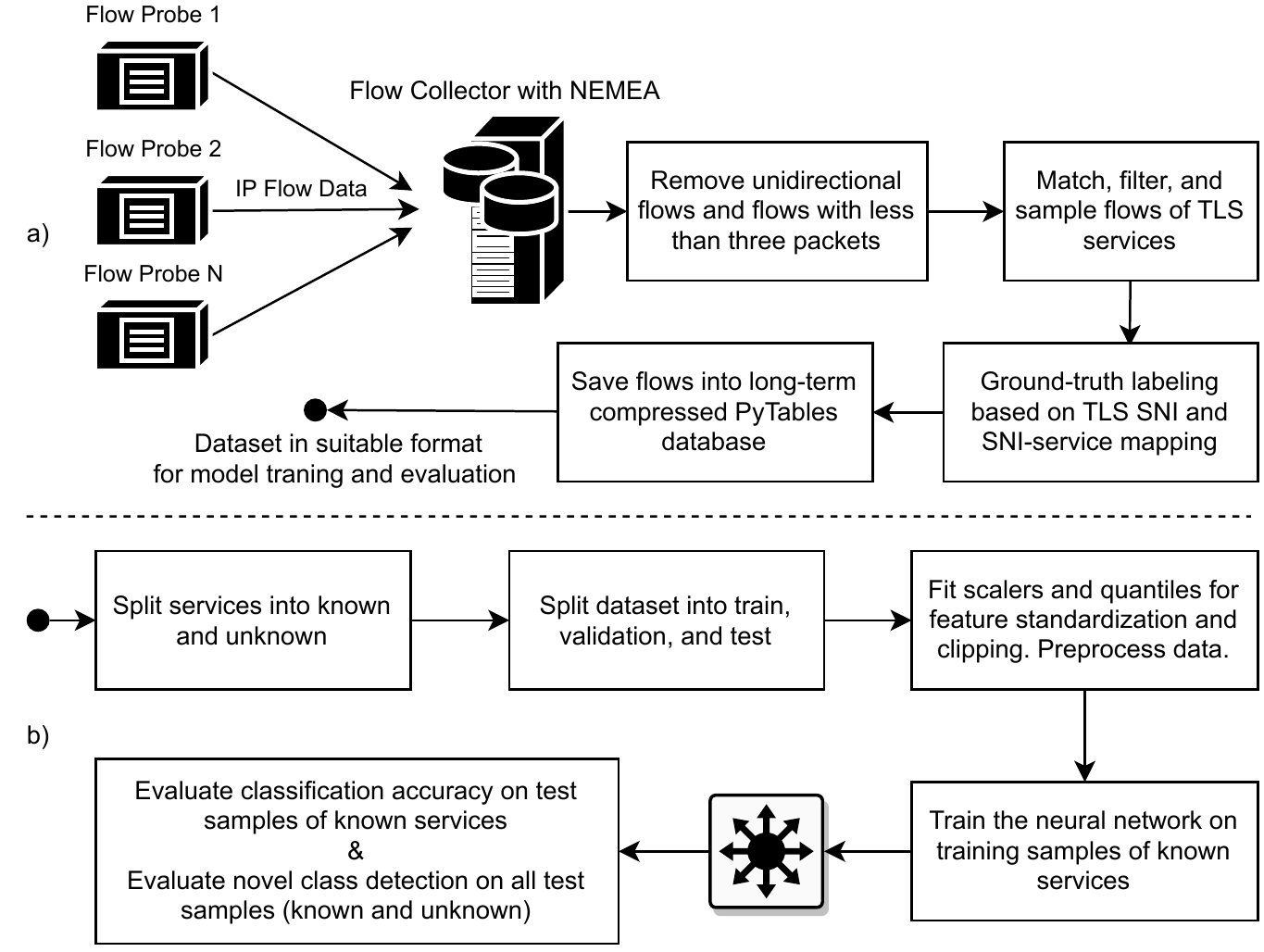}
	\caption{Visualization of a) the dataset collection process, b) the workflow for experiment evaluation.}
	\label{fig:data-processing}
\end{figure}

\section{Experimental setup}
\label{sec:experimental-setup}
We conduct our experiments on the MetaCentrum computing grid, on machines with 2 GPUs Nvidia Tesla T4 16\,GB, 32 CPUs Intel Xeon Gold 5218 2.30\,GHz, and 192\,GiB of RAM. The software stack for training, pre-processing, and visualizations includes PyTorch 1.9, Scikit-learn, TensorBoard, and PyTables as a database. We chose PyTables (which is based on the HDF5 format) over Pandas dataframes because of its built-in blosc\footnote{\url{https://www.blosc.org/}} compression (the dataset still has over 35\,GB) and fast random access, which is practical for training with random subsets of data.

The dataset spans two consecutive weeks of traffic. We split the dataset such that the samples of the first week are used for training and validation, and the second week is used as a test. Due to the asymmetric routing problem described in Section \ref{asymmetric-routing}, some of the dataset services do not have enough samples. We omitted services with less than 100 samples, which left us with 191 services (out of 200 labeled services in the dataset) for evaluation.

To account for randomness in network weight initialization, data bootstrapping, and dropout layers, we perform 10-fold cross-validation, with each fold having its own train-validation split of the first-week traffic. The second-week traffic is used as a test for all folds and is never used during training. Metrics presented in the Evaluation Section are averages across ten folds.

Our training pipeline follows standard practices with those notable choices. As an optimizer, we use AdamW, which improves the handling of weight decay over standard Adam. We experimented with cyclic learning rate schedulers; the best results were achieved with CyclicLR with a cycle length of 10 epochs. An advantage over the popular one-cycle approach is the training is less sensitive to the correct choice of the number of epochs. We train the neural network for 60 epochs; for each epoch, we sample a random subset of 500,000 training flows. We tested the following loss functions: standard cross-entropy, cross-entropy with label smoothing, Focal loss \cite{lin_FocalLoss_2017} (gives more focus on hard examples), and SimLoss \cite{kobs_SimLossClass_2020} (uses the hierarchy of classes). The standard cross-entropy turned out to be the best option. 

\section{Evaluation}
\label{sec:evaluation}
The goals of our experimental evaluation are threefold: (1) To show that the dataset is relevant and useful for the task of service identification with the reject option in TLS encrypted traffic. (2) To evaluate the model architecture, the influence of input data types (PSTATS vs FLOWSTATS), the influence of PSTATS features (packet sizes, times, and directions), and the importance of packet positions in PSTATS. (3)~To compare the novel class detection techniques using the created dataset.

The rest of this section describes the used performance metrics and the evaluation scheme for the detection of unknown services. Results are presented in Section~\ref{sec:results}.

\subsection{Classification metrics}
To evaluate the classification performance, we have measured accuracy, superclass accuracy, macro averaged F1-score, and its superclass version. Superclass metrics consider a prediction to be correct when the predicted superclass---the service group such as Google, Mozilla, ESET---is correct. Superclass metrics reflect that those wrong predictions among the same service group are less problematic.

\begin{figure}[h]
	\centering
	\includegraphics[width=0.58\linewidth]{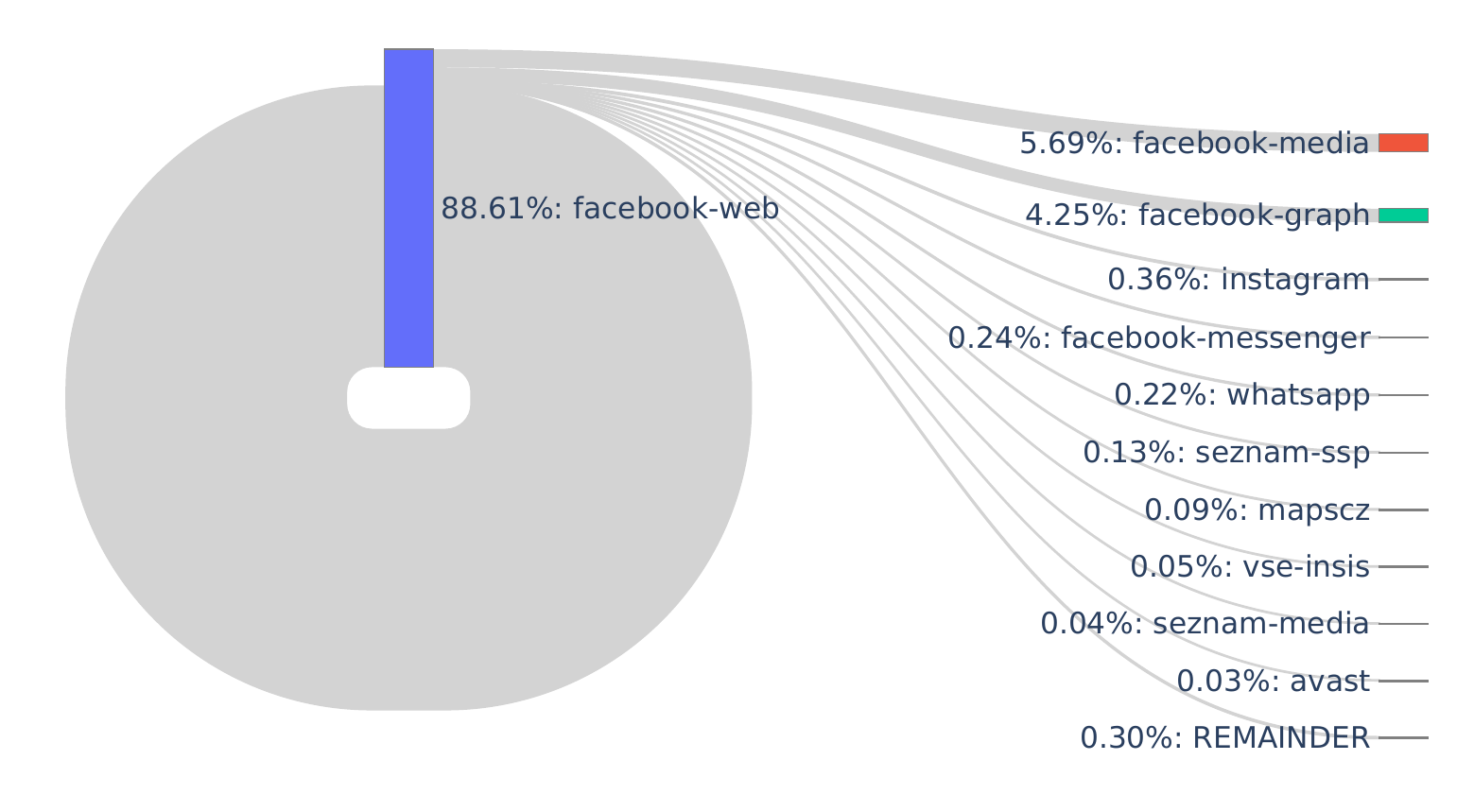}
	\caption{The distribution of predictions of the facebook-web service visualized in a Sankey plot.}
	\label{fig:sankey}
\end{figure}

We also visualize per-class Sankey plots showing the distribution of predictions. An example is shown in Figure \ref{fig:sankey}. It shows that 88.61\% of facebook-web predictions are correct; the rest (11.39\%) are misclassified samples where the classifier predicted a wrong class. The top five predicted services out of the mistakes belong to the correct superclass (facebook-media, facebook-graph, instagram, facebook-messenger, and whatsapp are provided by Facebook). Therefore, those can still be considered as correct predictions under superclass accuracy or superclass F1-score.

\subsection{Novel class detection metrics}
\label{sec:novelty-detection-metrics}
To evaluate the novel class detection performance, we have opted for two standard metrics: True Positive Rate (TPR) at a fixed 5\% False Positive Rate (FPR), which will be denoted as TPR@5\%FPR, and the area under the ROC curve (AUROC). Novel class detection can be seen and evaluated as binary classification: the novel, unknown class is positive, the normal class is negative. As explained in Section \ref{sec:reject-option}, the detection methods require a setting of the decision threshold. The threshold is computed over the validation set as the 95th percentile of the \textit{novelty scores}. TPR@5\%FPR measures the performance at a single decision point. On the other hand, AUROC measures performance over all possible decision threshold values. It is common to compute the area under the entire ROC. However, for the use-case of high-speed network monitoring, models with high FPR are not deployable in practice because a high volume of false positives leads to alert fatigue. Thus, we compute the AUROC just for the region between 0\% and 10\% FPR (the leftmost tenth of the curve). The AUROC is calculated with Scikit-learn function \texttt{roc\_auc\_score(max\_fpr=0.1)}.

The decision threshold is calculated over the validation set, which comes from the first-week traffic. When this threshold is used for the traffic of the second week (which can be seen as ``future'' during the testing), the intended setting of 5\% FPR does not hold. This is because the characteristics of the services' traffic change in time. As a result, more flows of known services receive higher \textit{novelty scores} and are classified as \textit{unknown}. Therefore, when we use the decision threshold to evaluate NC detection on the second-week traffic, we get $\sim$6.5\% FPR instead of the intended 5\% FPR. The takeaway is that traffic patterns and behavior of TLS services are not static, and a model like ours would require periodic re-calibration of decision thresholds to maintain the desired level of FPR. In Table \ref{tab:unknown-services-results} with NC detection results, we report both TPR@6.5\%FPR and adjusted TPR@5\%FPR. AUROC metric is not affected as it does not require a decision threshold.

\begin{table}[t]
    \small
	\centering
	\setlength\extrarowheight{1pt}
	\caption{The selected values of hyperparameters. Robust means scaling with median and inter-quartile range, standard means scaling with mean and variance.}
	\label{tab:hyperparameters}
	\begin{tabular}{| l | l  | r |}
		\hline
		\textbf{Hyperparameter} & \textbf{Scope} & \textbf{Value} \\ \hline
		Softmax temperature & softmax, gradient-based NC detection & 3 \\ \hline
		p-norm for per-sample gradients & gradient-based NC detection & 1.5 \\ \hline
		SimLoss alpha & gradient-based NC detection & 0.075 \\ \hline
		FLOWSTATS quantile clip & both NC detection and classification & 0.95 \\ \hline
		FLOWSTATS scaler & both NC detection and classification & robust \\ \hline
		PSTATS IAT min max clip &both NC detection and classification & 1ms, 15000ms \\ \hline
		PSTATS scaler - IAT & both NC detection and classification & standard \\ \hline
		PSTATS scaler - sizes & both NC detection and classification & standard \\ \hline
	\end{tabular}
\end{table}

\subsection{Evaluation scheme for unknown services detection}
To evaluate the detection of novel services, we have to split the dataset into known and unknown services, train the model on the known traffic, and evaluate on unknown. We selected three splitting points to measure performance for different numbers of known classes and ratios of known/unknown traffic. The splits are taking top-50, top-100, and top-150 services as known, and the rest (out of the 191  services) as unknown. Traffic samples of the services from the same group are put either all into known or all into unknown to avoid an unrealistic scenario of trying to detect, for example, some unknown Mozilla service while having a model trained on other Mozilla services. The evaluation workflow is visualized in Figure \ref{fig:data-processing}. We compare traffic characteristics of the known and unknown flows for the top-100 split and found no major differences, which would allow for a simple distinguishing of unknown flows and thus make the task easier. The comparison is shown in Figure~\ref{fig:dataset-stats-known-unknown}.

\section{Results}
\label{sec:results}
Table \ref{tab:hyperparameters} presents the values of hyperparameters, which were selected based on the 10-cross-validated performance for top-100 known services. The same parameters were reused for the evaluation of top-50, top-100, and top-150 known services. Using those parameters and the model described in Section \ref{sec:model-architecture}, we have achieved the classification results presented in Section \ref{sec:classification-results} and the unknown services detection results presented in Section \ref{sec:unknown-services-results}. Section~\ref{sec:data-input-impact} studies the impact of individual input features and their preprocessing, and Section~\ref{sec:model-interpretation} shows XAI interpretations of the trained neural network.

\subsection{Identification of known services}
\label{sec:classification-results}
The results in Table~\ref{tab:classification-results} show that the model is well capable of identifying services which it was trained on. Even for the hardest 191-classes case, over 97\% of test samples are correctly classified and over 98.5\% samples have a correct superclass prediction. The difference between the classification of 50 classes and 191 classes is 0.4\% of accuracy; a more noticeable 2\% difference can be observed when comparing achieved F1-scores and a 4\% difference when comparing superclass F1-scores. This can be attributed to the high imbalance in the number of samples per service. F1-score in a multi-class setting is computed as an average of F1-scores of individual services, with each service having an equal share regardless of the number of its samples. Therefore, adding more services with fewer samples (i.e., ``making the problem harder'') does influence the achieved F1-score more than it does accuracy. As a result, F1-score is more suitable for comparison of classification performance in high-imbalance settings, such as the one with web services traffic. Detailed per-class results for the 100-classes case are shown in a classification report in~\ref{appendix:classification-report}.

\begin{table}[t]
    \small
	\centering
	\setlength\extrarowheight{1pt}
	\caption{Classification results without reject option for top-50, top-100, top-150, and all classes. Results are shown in the format average (\textpm std.) obtained over ten folds.}
	\label{tab:classification-results}
	\begin{tabular}{| l | c  | c | c | c |}
		\hline
		\textbf{\#Classes} & \textbf{Accuracy} & \textbf{Superclass Accuracy} & \textbf{F1-score} & \textbf{Superclass F1-score} \\ \hline
	    50 & 97.44 (\textpm 0.04) & 99.42 (\textpm 0.01) & 96.14 (\textpm 0.05) & 99.49 (\textpm 0.01) \\ \hline	    
	    100 & 97.41 (\textpm 0.04) & 99.07 (\textpm 0.02) & 95.91 (\textpm 0.06) & 97.85 (\textpm 0.03) \\ \hline
	    150 & 97.17 (\textpm 0.03) & 98.79 (\textpm 0.02) & 95.60 (\textpm 0.08) & 97.20 (\textpm 0.06) \\ \hline
		all 191 & \textbf{97.04 (\textpm 0.05)} & \textbf{98.65 (\textpm 0.03)} & \textbf{94.20 (\textpm 0.11)} & \textbf{95.49 (\textpm 0.10)} \\ \hline
	\end{tabular}
\end{table}

\subsection{Comparison to ML model}
\label{sec:ML-DL-comparison}
Based on the discussion in Section~\ref{sec:model}, we compare the classification performance of our neural network with a standard ML model. We choose the LightGBM model, which is a gradient boosting ensemble of decision trees. We use the same input data for training the ML model as we do for the neural network; no additional features were computed. Yang~et~al.~\cite{yang_DeepLearning_2021} performed a similar experiment and were able to tune a tree-based model to exceed the encrypted traffic classification performance of a convolutional neural network. In our experiments, we were not able to achieve that, and our neural network provides better classification performance (apart from its advantages in unknown traffic detection). The comparison is summarized in Table~\ref{tab:ML-DL-comparison}. The ML model achieved commendable 96.63\% superclass accuracy, which is only 2.44\% worse than our neural network. However, the differences are larger in other metrics; for example, there is almost a 5.82\% gap in F1-score. We have chosen a neural network as our model for its advantages in unknown traffic detection, but the improvement in classification performance is, of course, a welcomed bonus.

% We believe that the LightGBM model can be further tuned to improve its performance to some extent

\begin{table}[t]
    \small
	\centering
	\setlength\extrarowheight{1pt}
	\caption{A classification performance comparison between DL and ML models. The results for DL are taken from Table~\ref{tab:classification-results}, the row with 100 classes. Results for LightGBM are averages across three folds. LightGBM was trained with those hyperparameters: \texttt{min\_data\_in\_leaf} 20, \texttt{feature\_fraction} 0.9, \texttt{bagging\_fraction} 0.8, \texttt{bagging\_freq} 1, \texttt{early\_stopping\_rounds} 10.}
	\label{tab:ML-DL-comparison}
	\begin{tabular}{| l | c  | c | c | c |}
		\hline
		\textbf{Model} & \textbf{Accuracy} & \textbf{Superclass Accuracy} & \textbf{F1-score} & \textbf{Superclass F1-score} \\ \hline
		LightGBM & 93.32 & 96.63 & 90.09 & 93.50 \\ \hline
		DL & 97.41 \textbf{(+4.09)} & 99.07 \textbf{(+2.44)} & 95.91 \textbf{(+5.82)} & 97.85 \textbf{(+4.35)}
		\\ \hline
	\end{tabular}
\end{table}

\subsection{Detection of unknown services}
\label{sec:unknown-services-results}
The novel class detection results are presented in Table~\ref{tab:unknown-services-results}. The first observation is that there are no significant differences between the performance of the top-50, top-100, and top-150 evaluations of the gradient-based unknown services detection. The TPR@5\%FPR values of the gradient-based method show that the achieved performance remains almost the same. Looking at the TPR@5\%FPR values of the energy-based method, we can see a pattern of the NC detection problem being harder for more known classes (as is the case for classification). The second observation is that the variance of measured NC detection metrics (shown in the form of standard deviation) is much higher than the variance of classification metrics. The neural network is not directly trained to detect unknown services; in other words, the loss function changes the network weights to improve classification performance, not NC detection performance. This makes the NC detection metrics more affected by the randomness of the training process.

\begin{figure}[t]
	\centering
	\includegraphics[width=0.7\linewidth]{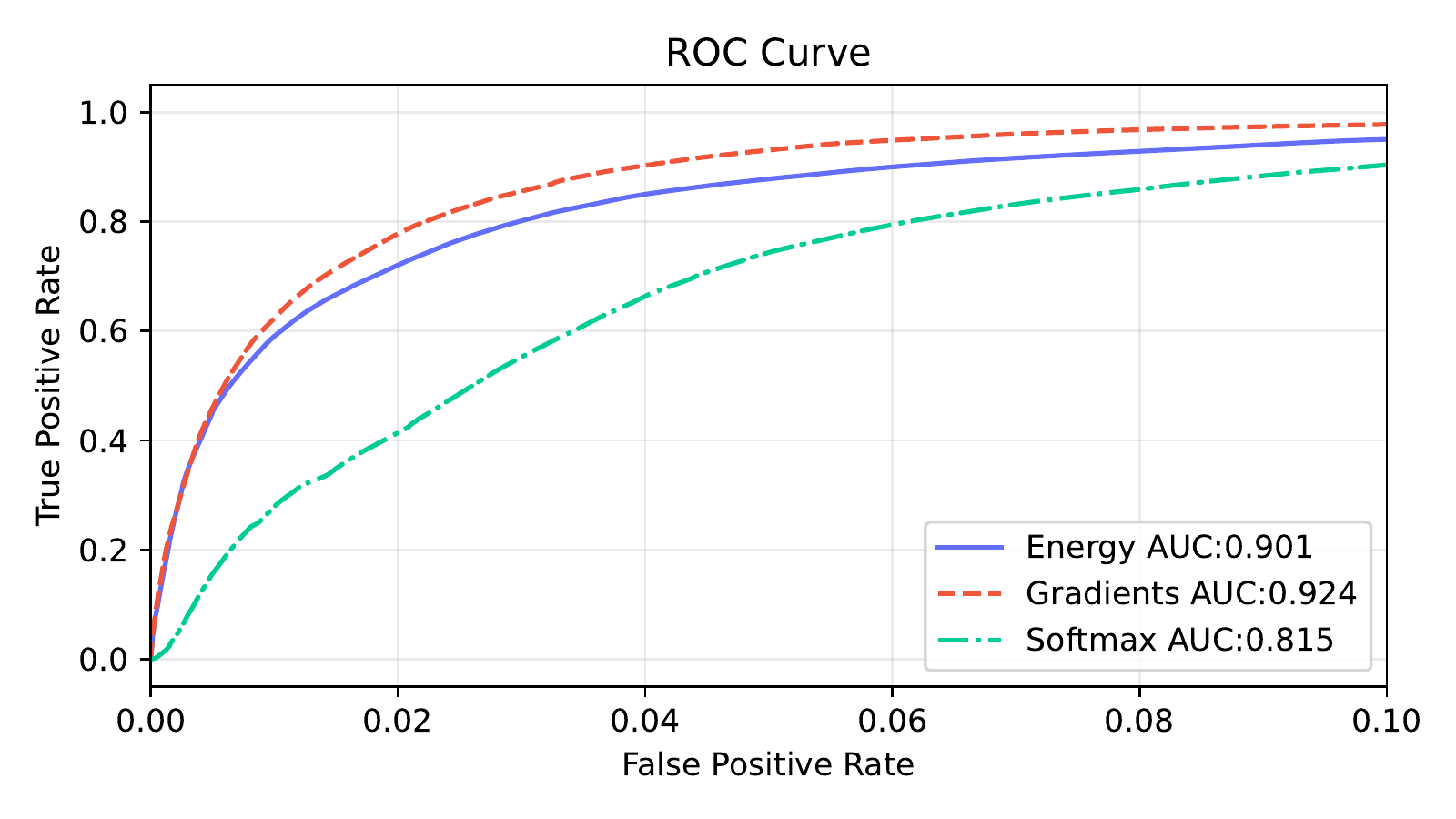}
	\caption{An ROC curve of the unknown services detection performance for top-100 known services and 91 unknown services. The ROC is shown for the region between 0 and 0.1 FPR. AUROC (AUC in the figure's legend) is computed over this region as described in Section~\ref{sec:novelty-detection-metrics}.}
	\label{fig:roc}
\end{figure}

We choose the top-100 case for further discussion. The best results for the detection of unknown services were achieved with the gradient-based method, which has 91.94\% TPR@5\%FPR and 92.13\% AUROC. The improvement over the energy-based method is 4\% in TPR and 2\% in AUROC. A comparison in the form of the ROC curve is shown in Figure~\ref{fig:roc}. The energy-based method achieved almost 13\% improvement in TPR over the softmax score baseline. Because the energy-based method is almost as fast as the softmax score baseline and has a trivial implementation, we believe that \textit{it can be considered as the new baseline method in future research works}.

\begin{table}[t]
    \small
	\centering
	\setlength\extrarowheight{1pt}
	\caption{Novel class detection results for top-50/141, top-100/91, and top-150/41 known/unknown split points. Column \#C indicates the number of top-N known classes. Results are shown in the format average (\textpm std.) obtained over ten folds. TPR is shown for 6.5\% FPR and 5\% FPR as discussed in Section \ref{sec:novelty-detection-metrics}.}
	\label{tab:unknown-services-results}
	\setlength{\tabcolsep}{.83\tabcolsep}
	\begin{tabular}{| l | c  | c | c | c | c | c |}
		\hline
		\multirow{2}{*}{\textbf{\#C}} & \multicolumn{3}{c|}{\textbf{TPR @6.5\%FPR and @5\%FPR}} & \multicolumn{3}{c|}{\textbf{AUROC}} \\ \cline{2-7}
		 & \textbf{Softmax} & \textbf{Energy} & \textbf{Gradients} & \textbf{Softmax} & \textbf{Energy} & \textbf{Gradients} \\ \hline
        50 & \makecell{79.62 (\textpm 0.99) \\ 75.01 (\textpm 1.12)} & \makecell{91.30 (\textpm 0.73) \\ 89.21 (\textpm 0.88)} & \makecell{93.46 (\textpm 0.61) \\ 91.86 (\textpm 0.64)} & 82.29 (\textpm 0.39) & 91.39 (\textpm 0.47) & 92.62 (\textpm 0.91)\\ \hline		
        100 & \makecell{79.98 (\textpm 1.33) \\ 74.97 (\textpm 1.28)} & \makecell{90.38 (\textpm 0.88) \\ \textbf{87.83 (\textpm 0.89)}} & \makecell{94.02 (\textpm 0.59) \\ \textbf{91.94 (\textpm 0.65)}} & 82.01 (\textpm 0.54) & \textbf{90.29 (\textpm 0.59)} & \textbf{92.13 (\textpm 0.28)}\\ \hline
		150 & \makecell{80.01 (\textpm 2.77) \\ 75.49 (\textpm 2.85)} & \makecell{89.98 (\textpm 1.37) \\ 86.73 (\textpm 1.38)} & \makecell{94.71 (\textpm 0.58) \\ 92.06 (\textpm 0.88)} & 83.01 (\textpm 1.09) & 89.20 (\textpm 0.63) & 91.98 (\textpm 0.28)\\ \hline
	\end{tabular}
\end{table}

\subsection{Data input and preprocessing impact}
\label{sec:data-input-impact}
A simple ablation process is used to find out how the model would perform if a data input or its preprocessing was modified or removed. Even though this is not the main focus of this work, we include a brief evaluation of different data types in the dataset (FLOWSTATS, packet times, packet directions) and assess their usefulness for the tasks of encrypted traffic analysis. We measure both classification accuracy and NC detection TPR@5\%FPR (for the top-100 classes split) to find out whether the modification impacts both tasks. We used the same hyperparameters values as listed in Table \ref{tab:hyperparameters}.

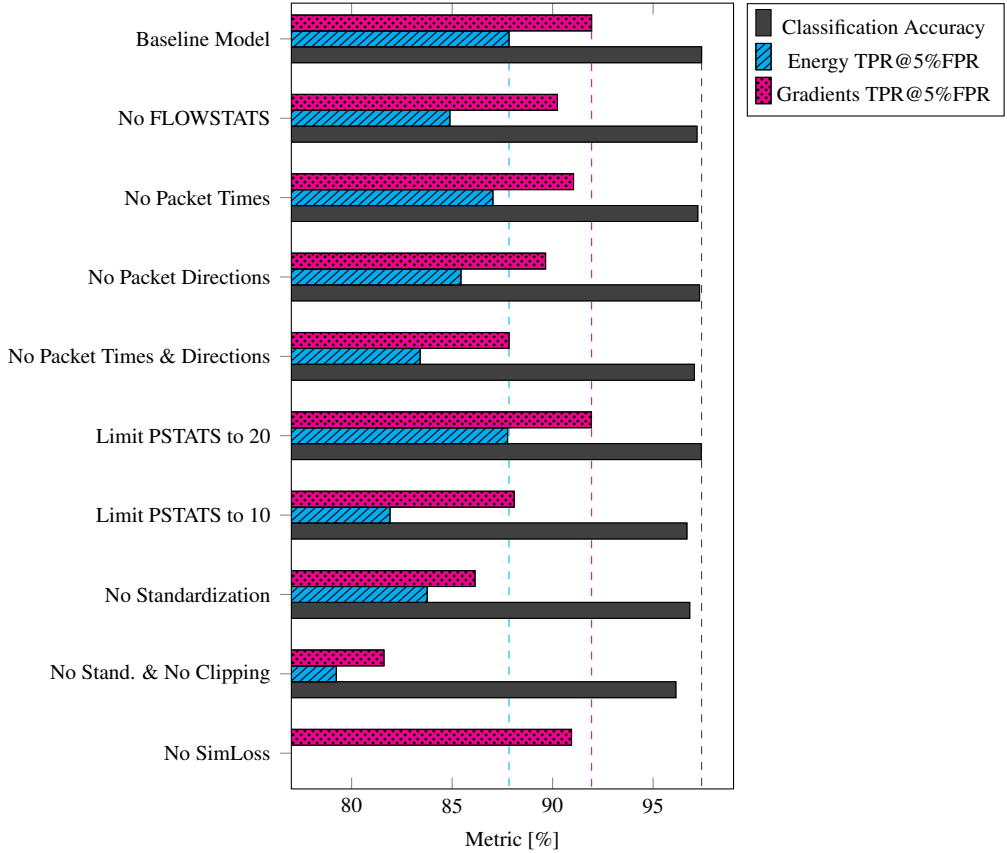
\begin{figure}[t]
\centering
\begin{tikzpicture}
\begin{axis}[
    tick label style={font=\footnotesize},
    label style={font=\footnotesize},
    width=7.4cm,
    height=12cm, 
    enlarge y limits=0.05,
    enlarge x limits=0.05,
    ylabel={},
    xlabel={Metric [\%]},
    symbolic y coords={
        Baseline Model,
        No FLOWSTATS,
        No Packet Times,
        No Packet Directions,
        No Packet Times \& Directions,
        Limit PSTATS to 20,
        Limit PSTATS to 10,
        No Standardization,
        % No Clipping,
        No Stand. \& No Clipping,
        No SimLoss,
    },
    ytick=data,
    y dir=reverse,
    ytick pos=left,
    xbar=0pt,
    xmax=98,
    xmin=78,
    bar width=6pt,
    legend image code/.code={
        \draw [#1] (0cm,-0.1cm) rectangle (0.2cm,0.25cm);
    },
    legend pos=outer north east,
    legend style={font=\footnotesize},
    ]
    % \tikzset{every node}=[font=\sffamily]
    \addplot [black, fill=darkgray, forget plot] coordinates {
        (97.41,Baseline Model)
        (97.18,No FLOWSTATS)
        (97.22,No Packet Times)
        (97.31,No Packet Directions)
        (97.05,No Packet Times \& Directions)
        (97.39,Limit PSTATS to 20)
        (96.68,Limit PSTATS to 10)
        (96.82,No Standardization)
        % (97.40,No Clipping)
        (96.13,No Stand. \& No Clipping)
        };
    \addplot[darkgray, dashed,line legend,sharp plot,nodes near coords={}, forget plot,
    update limits=false,shorten >=-10mm,shorten <=1mm] coordinates {(97.41,Baseline Model) (97.41,No SimLoss)};
    \addplot[magenta, dashed,line legend,sharp plot,nodes near coords={}, forget plot,
    update limits=false,shorten >=-10mm,shorten <=-1mm] coordinates {(91.94,Baseline Model) (91.94,No SimLoss)};
    \addplot[cyan, dashed,line legend,sharp plot,nodes near coords={}, forget plot,
    update limits=false,shorten >=-10mm,shorten <=+1 mm] coordinates {(87.83,Baseline Model) (87.83,No SimLoss)};
    \addplot [black, fill=darkgray] coordinates {
        (97.41,Baseline Model)
        (97.18,No FLOWSTATS)
        (97.22,No Packet Times)
        (97.31,No Packet Directions)
        (97.05,No Packet Times \& Directions)
        (97.39,Limit PSTATS to 20)
        (96.68,Limit PSTATS to 10)
        (96.82,No Standardization)
        % (97.40,No Clipping)
        (96.13,No Stand. \& No Clipping)
        };
    \addplot [black, fill=cyan, postaction={pattern=south west lines}] coordinates {
        (87.83,Baseline Model)
        (84.89,No FLOWSTATS)
        (87.03,No Packet Times)
        (85.44,No Packet Directions)
        (83.41,No Packet Times \& Directions)
        (87.76,Limit PSTATS to 20)
        (81.91,Limit PSTATS to 10)
        (83.76,No Standardization)
        % (87.79,No Clipping)
        (79.24,No Stand. \& No Clipping)
        };
    \addplot [black, fill=magenta, postaction={pattern=crosshatch dots}] coordinates {
        (91.94,Baseline Model)
        (90.24,No FLOWSTATS)
        (91.04,No Packet Times)
        (89.65,No Packet Directions)
        (87.83,No Packet Times \& Directions)
        (91.93,Limit PSTATS to 20)
        (88.09,Limit PSTATS to 10)
        (86.14,No Standardization)
        % (91.91,No Clipping)
        (81.61,No Stand. \& No Clipping)
        (90.94,No SimLoss)
    };
    \legend{Classification Accuracy, Energy TPR@5\%FPR, Gradients TPR@5\%FPR}
\end{axis}
\end{tikzpicture}
%%%
	\caption{Evaluation of input data types, preprocessing steps, and SimLoss. The results from Tables \ref{tab:classification-results} and \ref{tab:unknown-services-results} are considered as the Baseline Model.}
	\label{fig:impact}
\end{figure}

\medskip\noindent
We evaluate these modifications:
\begin{itemize}
    \item Remove FLOWSTATS features
    \item Remove packet direction and/or packet times, but keep packet sizes
    \item Limit the length of packet sequences to the size of 20 and 10
    \item Do not use feature standardization and/or clipping
    \item Use standard cross-entropy loss instead of SimLoss for gradient-based NC detection
\end{itemize}

The results are visualized in Figure \ref{fig:impact}. A surprising observation is that none of the modifications have a substantial impact on the classification performance. On the other hand, the impact on NC detection performance is considerable in most cases. The rest of this section discusses a couple of examples. A possible conclusion is that future research works elaborating on network traffic features should perform the evaluation on harder problems, such as the detection of unknown traffic, rather than on classification tasks, which might not be impacted that much.

\paragraph{Not using packet times and directions}
Even though not using packet times and packet directions has less than 0.5\% negative impact on accuracy, the decrease in TRP@5\%FPR is more than 4\% for both gradient and energy-based methods. 

\paragraph{Effect of feature standardization and clipping}
Omitting both feature standardization and clipping and thus exposing the model to the effect of outliers has a severe impact on NC detection; both the energy and gradient-based method suffer a more than 8\% drop in TPR while the classification accuracy decrease is 1.3\%. Using clipping without standardization has a similar but smaller effect. Using standardization without clipping (not shown in Figure~\ref{fig:impact}) achieved almost the same results as the baseline model. We leave further analysis of standardization and clipping of traffic features (packet sizes, packet times, flow statistics) and their combined effect on NC detection performance for future work.

\paragraph{Shorter PSTATS sequences}
Limiting PSTATS to the length of 20 has no effect on both tasks; limiting it to 10, however, has a substantial effect. The distribution of the number of packets in flows can explain this---75\% of dataset flows do not have over 20 packets and thus are not affected when the PSTATS length is limited. 

\paragraph{SimLoss}
In Section \ref{sec:gradient-based}, we have introduced a modification of the original gradient-based method. The idea was to replace the loss function used for gradient computation with SimLoss. We evaluated the impact of using SimLoss and measured a 1\% increase in TRP@5\%FPR and a 0.6\% increase in AUROC. Using SimLoss allows similarities between individual classes to be defined. The trained model is expected to make more mistakes between those classes, i.e., the model is expected to be unsure (low softmax score of the predicted class). When SimLoss is used for gradient-based NC detection, the a priori knowledge of similar classes is leveraged to produce fewer false positives among classes that are expected to be mismatched. We consider this original idea to be worth elaborating on in future work.

\subsection{Model interpretation}
\label{sec:model-interpretation}
Neural networks are known to be hard to interpret as opposed to standard ML models, such as tree-based models. The understanding of models' inner workings is crucial to assess their correct functioning, and detailed explanations for predictions are important for practical deployments, where classification models are part of bigger analysis systems. To address this need, a field of explainable AI (XAI) has emerged~\cite{barredoarrieta_ExplainableArtificial_2020}.

\begin{figure}[t]
\centering  
\begin{subfigure}[b]{1\textwidth}
    \centering 
    \includegraphics[width=0.81\linewidth]{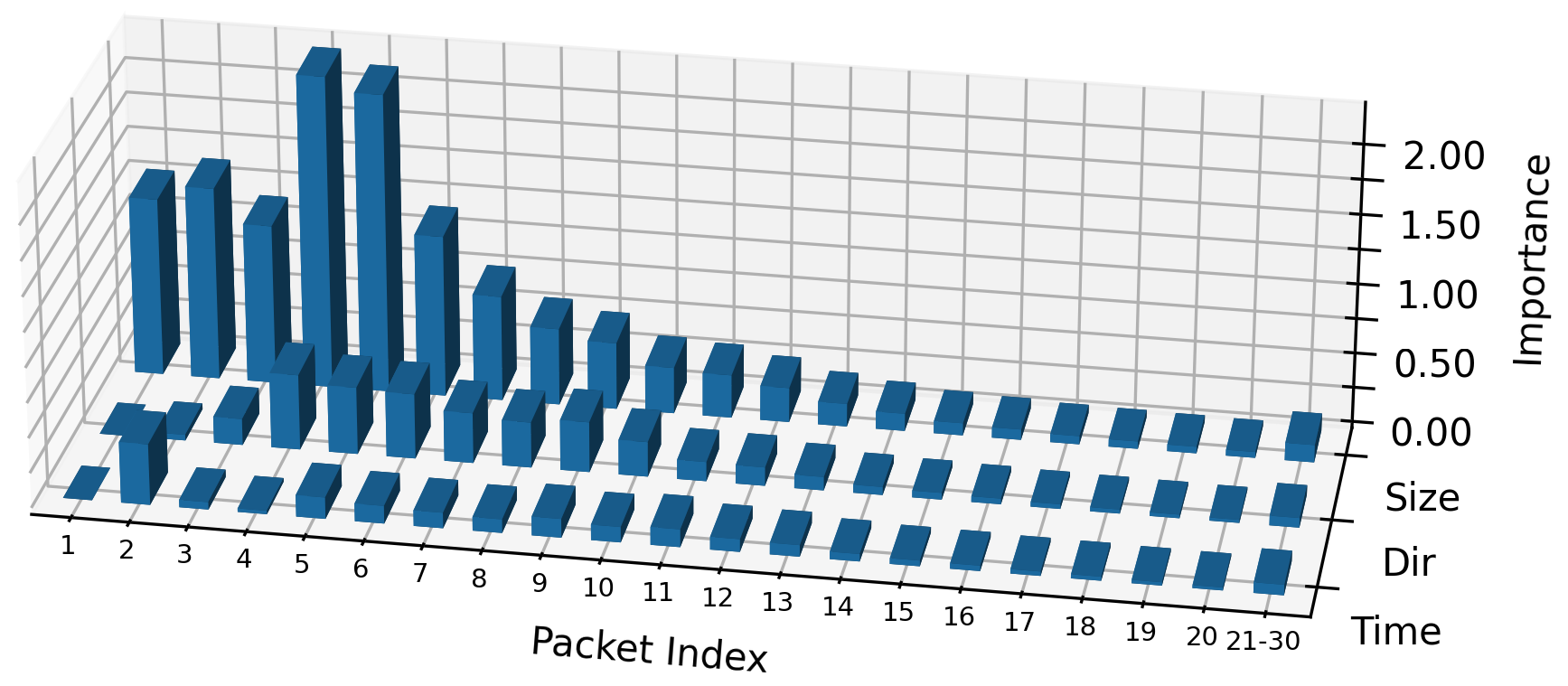}
    \caption{XAI interpretations of packet metadata sequences. As expected, packet sizes are the most important data input for model decisions. What also is not surprising is that packets exchanged at the start of the communication are more important than later packets. The first couple of packets (the exact number depends on the TLS version, session resumption, the size of the server certificate, etc.) carry TLS messages for the TLS handshake. We interpret the peak around the 5th and 6th packets as related to the first application data exchange after the handshake is completed. However, the first three packets, which will always carry TLS messages, also have high importance; our interpretation is that the sizes of ClientHello messages and the sizes of server certificates are still useful for making model predictions.}
    \label{fig:xai-pstats} 
\end{subfigure}

% \\

\begin{subfigure}[b]{1\textwidth}
    \centering  
    \includegraphics[width=0.60\linewidth]{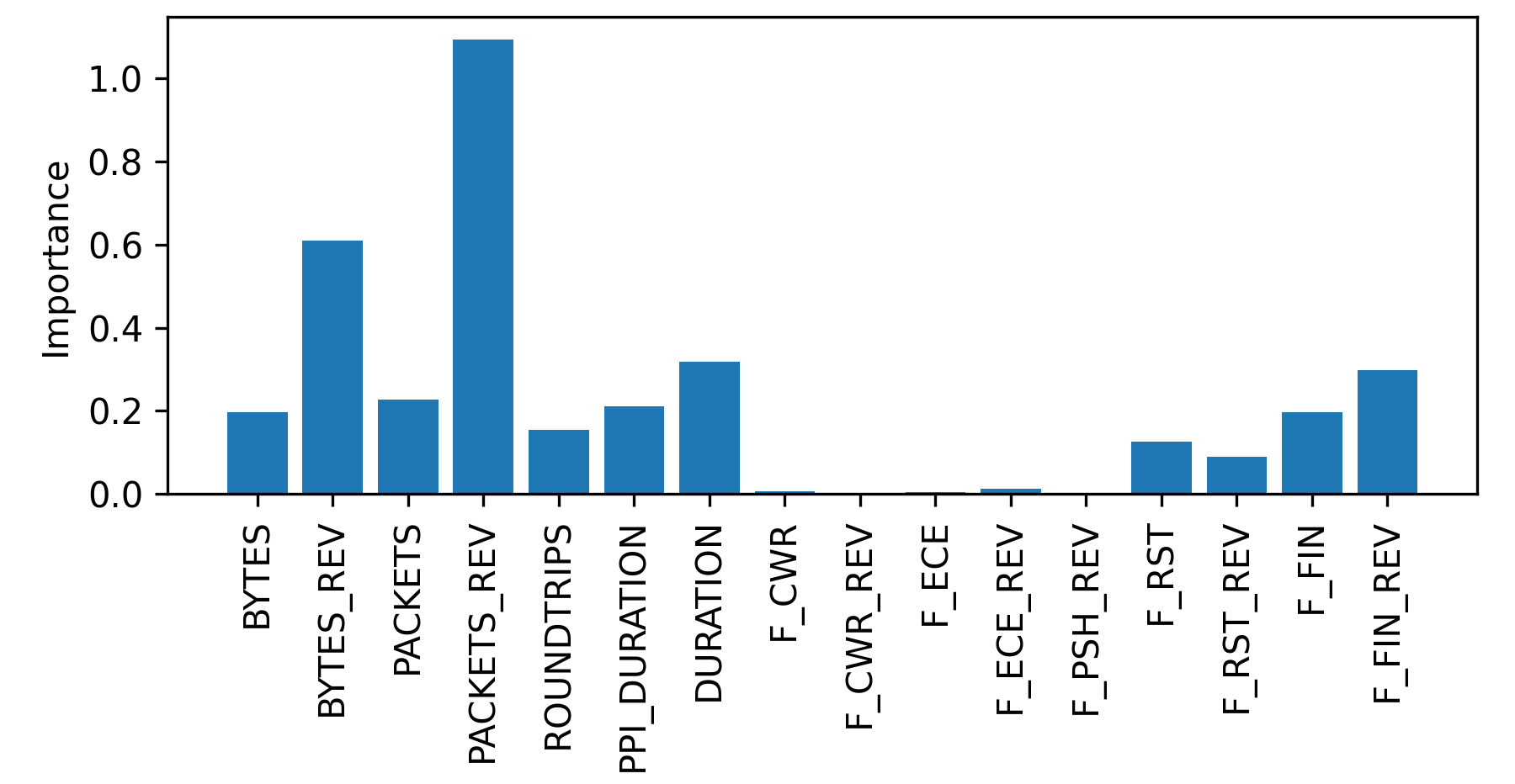}
    \caption{XAI interpretations of flow statistics. Feature names ending with \texttt{\_REV} describe the reverse communication direction (server to client). The \texttt{PPI\_DURATION} feature measures the time duration of the PSTATS data, while \texttt{DURATION} measures the entire flow. Feature names starting with \texttt{F\_} describe the presence of individual TCP flags. The most important features are the number of packets and bytes in the reverse direction, the flow duration, and the presence of TCP FIN flags. The TCP FIN flag is used for the termination of a TCP connection, and the model could be using this information to determine whether the connection was closed correctly.}
    \label{fig:xai-flowstats}
\end{subfigure}

\caption{XAI interpretations of our neural network. Importances are computed as averages of absolute values of SHAP values as described in Section~\ref{sec:model-interpretation}. The scale of the y \textit{Importance} axis is the same for both figures, and values can be compared across both figures. For example, the packet size at the 6th position is as important as the \texttt{PACKET\_REV} feature (the number of packets in the reverse communication direction).}
\label{fig:xai}
\end{figure}

In this section, we provide a short interpretation analysis of our neural network. We have used a popular XAI method called SHapley Additive exPlanations (SHAP)~\cite{lundberg_UnifiedApproach_2017}. The core idea is to transform the model such that its predictions can be computed from input features in an additive fashion (the transformed additive features are called attributions or SHAP values\footnote{An important detail for computing SHAP values is the selection of a baseline(s). For image processing, a black image is often used as a ``no-signal'' reference. For our data type, however, there is no such natural reference. We have followed a recommendation of the SHAP method author and used 20 weighted k-medians clusters as the reference baselines. \url{https://github.com/slundberg/shap/issues/23}.}). Global model explanations are then computed as an aggregation of local (per-sample) attributions. A recent related paper is by Nascita~et~al.~\cite{nascita_XAImeets_2021}, which provides a comprehensive analysis of XAI methods for interpretation of encrypted traffic classifier. We chose a similar approach as in this related work and compute the global explanations as averages of absolute values of samples attributions. 

As opposed to the analysis presented in the previous section, where the impact of individual features and preprocessing steps is measured as the difference in a performance metric after model retraining, the SHAP method works with a single trained model. The SHAP interpretation of our neural network is presented in Figure~\ref{fig:xai}, which shows the global model importances of individual packet position and flow statistics. The importances in Figure~\ref{fig:xai} are computed as averages across samples of all services. To demonstrate that the model relies on different parts of input for different services, we show per-class importances in Figure~\ref{fig:xai-services}.

\section{Related work}
\label{sec:relatedwork}
The closest related work is \cite{shbair_multilevelframework_2016}, in which Shbair et al. targeted the same goal of identifying web services in TLS traffic. The proposed method is based on computed statistical features, the dataset was generated in a lab environment, and TLS SNI was used for ground-truth labeling. There is a two-level hierarchy of services and service providers, however, the differentiation between services is based on the prefix structure of the SNI domain name: service provider corresponds to a common second-level domain, each subdomain corresponds to a different service (though some domain preprocessing is done like removing dashes and numbers). Classification is two-level: one service provider classifier and one service classifier per provider. The best model was RandomForest, which achieved 93.1\% service identification accuracy. The main differences from our work are that (1) our dataset is built from real traffic, (2) as a model, we choose neural networks, which have advantages for implementing the detection of unknown services (which is not handled in this related work), (3) we carefully selected services and corresponding SNIs and did not rely on prefix structure for differentiation between services (for example, our service YouTube has among others following domains: *.youtube.com, *.youtube.googleapis.com, *.ytimg.com), nor between providers (we consider Facebook, Instagram, and WhatsApp to have the same provider even though their top-level domains are different). We believe that the prefix structure of domain names cannot nowadays be used for service differentiation because it is too common that services have multiple top-level domains or use different domains for sub-functionalities.

Recent influential work by Yang et al. is \cite{yang_DeepLearning_2021}, which also targeted application identification in network traffic (not limited to the TLS protocol) with ``zero'' traffic detection, i.e., detection of the traffic that is unknown at the time of training. Their dataset is built with a commercial firewall appliance, which is able to provide ground truth labels for hundreds of applications. The evaluation is performed on an admirable amount of 835 applications. The proposed method, which is based on 1D convolutions and gradient-based NC detection, achieved 91\% classification accuracy for 200 classes and 78\% TPR at 1\% FPR for unknown TCP traffic detection with 162+500 known+unknown classes. In comparison to these results, we were not able to achieve that high detection rate of unknown services (our best result for 1\% FPR is 66.8\% TPR); on the other hand, we achieved higher classification accuracy with a model of similar size (ours has 1.1 million, theirs 1.4 million parameters). However, because we used a different dataset with different services, the results are not suitable for direct comparison.

Due to the legal and business aspects, their dataset was not made public. This motivated us to build a similar dataset (though limited to TLS traffic) with as many application labels as possible and share it openly with the traffic classification research community. This is in alignment with one of the concluding remarks in  \cite{yang_DeepLearning_2021} claiming that \textit{``constructing an open corpus with rich class diversity and large class cardinally K should be a priority goal to allow for meaningful and fair cross-comparison of research proposals.''} We consider CESNET2, the national research and education network infrastructure, to be in an ideal position to join the effort to build large, diverse, and public datasets of labeled network traffic.

Another recent work by Akbari et al. is \cite{akbari_LookCurtain_2021} targeting service classification in encrypted TLS traffic. Their public dataset Orange'20 is built from the traffic of a major mobile ISP, has 19 application classes, 8 traffic categories, and 120k of labeled flows. The dataset was used for training a neural network with a multi-modal architecture, processing three types of data: packet time series, flow statistics, and raw payload bytes of TLS handshakes. The payload bytes are processed with 1D convolutions, however, some fields, such as TLS SNI and cipher information, are masked. TLS SNI has to be removed from the input data because it was used for ground-truth labeling. The packet sequences are processed with recurrent LSTM layers. Their approach achieved 97\% classification accuracy for the 19 classes; unknown traffic detection was not considered. Our main concern, which is also pointed out in \cite{yang_DeepLearning_2021}, is that the dataset's low number of application classes makes the classification problem easier to solve.

Shahraki et al. in \cite{shahraki_ActiveLearning_2021} point out the lack and the importance of public network traffic datasets. The authors analyzed the active learning approach for traffic classification and concluded that the lack of up-to-date datasets of encrypted traffic hinders the improvement of classification techniques and their evaluation. The Internet, its protocols, and web services are all evolving, and thus network traffic older than a couple of years might not be relevant anymore. For example, the most used public dataset for traffic classification, the ISCXVPN2016 dataset, is more than five years old, and the popular CTU-13 botnet traffic dataset is from 2011 and is still being used in new research works because there are no better public alternatives.

\begin{table}[t]
    \small
	\centering
    \begin{threeparttable}
	\setlength\extrarowheight{1pt}
	\caption{The overview of existing public labeled datasets for traffic classification.}
	\label{tab:datasets-overview}
	\begin{tabular}{| l | l  | l | r | r |}
		\hline
		\textbf{Dataset Name} & \textbf{Year} & \textbf{Scope} & \textbf{\#Flows} & \textbf{\#Classes} \\ \hline
		This work (CESNET-TLS22) & 2022 & TLS & 140M & 191\\ \hline
		Orange'20 \cite{akbari_LookCurtain_2021} & 2021 & TLS & 120k & 19\\ \hline
		MIRAGE Video \cite{montieri_Packetlevelprediction_2021} & 2021 & Mobile apps & 14k & 8\\ \hline
		UC Davis \cite{rezaei_HowAchieve_2020} & 2020 & QUIC & 6.5k & 5\\ \hline
		MIRAGE \cite{aceto_MIRAGEMobileapp_2019} & 2019 & Mobile apps ($\sim$80\% is TLS) & 97k & 41\\ \hline
		ISCX VPN-nonVPN \cite{draper-gil_CharacterizationEncrypted_2016} & 2016 & both TCP and UDP & 11.6k\tnote{a} & 17\\ \hline
	\end{tabular}
	\begin{tablenotes}
      \item[a]{\footnotesize This number is reported in \cite{aceto_DISTILLEREncrypted_2021} as the number of labeled flows after cleaning network broadcasts.}
    \end{tablenotes}
  \end{threeparttable}
\end{table}

\begin{table}[t]
    \small
	\centering
	\setlength\extrarowheight{1pt}
	\caption{Notable examples of datasets which are not public.}
	\label{tab:datasets-overview-nonpublic}
	\begin{tabular}{| l | l  | l | r | r |}
		\hline
		\textbf{Dataset Name} & \textbf{Year} & \textbf{Scope} & \textbf{\#Flows} & \textbf{\#Classes} \\ \hline
		Yang el al. \cite{yang_DeepLearning_2021} & 2021 & both TCP and UDP & 10.4M & 835\\ \hline
		Rezaei el al.\cite{rezaei_LargescaleMobile_2020} & 2019 & Mobile apps (40\% is TLS) & unclear & 80\\ \hline
		Aceto et al. \cite{aceto_MobileEncrypted_2019} & 2018 & Mobile apps & 121k & 94\\ \hline
		Lopez-Martin et al. \cite{lopez-martin_NetworkTraffic_2017} & 2017 & both TCP and UDP & 266k & 108\\ \hline
	\end{tabular}
\end{table}

\subsection{Available public datasets}
We surveyed available public datasets of labeled traffic; their overview is shown in Table \ref{tab:datasets-overview}. To our best knowledge, there are no other available works that would provide public datasets for encrypted traffic classification. For comparison, we also show a few examples of non-public datasets in Table \ref{tab:datasets-overview-nonpublic}. For each dataset, we provide the year it was created, its scope, the number of labeled flows, and the number of class labels.

\section{Conclusions}
Cyber security and operational processes in modern networks depend on information from network monitoring systems. In recent years, network monitoring research has benefited from technology progress of machine learning and artificial intelligence to deal with the challenge of encrypted traffic and retrieve useful knowledge about the situation in the network. Machine learning can improve the precision of critical processes such as network traffic analysis, classification, and threat detection. However, an essential prerequisite for those methods is high-quality and up-to-date datasets of real traffic with appropriate labels. The main goal of our work was to create and evaluate a suitable dataset for encrypted traffic analysis, since we agree with the state-of-the-art authors that there is a lack of recent datasets for this task. We have created a new dataset consisting of almost 200 TLS services using a combination of manual work and developed open-source tools, which can be reused to extend this dataset or to create a new dataset in a different environment. Guidelines on how to reproduce the dataset collection process together with a Dockerfile and example configuration files are available at the dataset download page.

To prove the usefulness of our dataset, we have trained a neural network with a multi-modal architecture and implemented three reject-option methods for the detection of unknown traffic. With the trained model, we achieved a classification accuracy of 97.04\% for individual services and 98.65\% for superclasses (groups of related services). One can argue that there is not much room for improvement in the classification performance; however, in the context of high-speed network monitoring, the processing speed is of the most importance, and we suggest future works can focus on optimization of the classification models for the inference speed, resource consumption, etc.

Opposed to the service classification, current methods for unknown traffic detection have yet to achieve satisfying results, making it an excellent area for future research. Our dataset can be leveraged to evaluate and compare improvements in this area. Existing public datasets are not large enough and do not contain enough classes to provide a suitable benchmarking problem. Compared to the existing public datasets of labeled network traffic, the number of flows in our dataset is three orders of magnitude higher, and the number of service labels, which is important to make the problem hard and realistic, is four times higher.

Using our dataset, we have compared three existing methods for the detection of unknown traffic. The best result of a 91.94\% detection rate with a 5\% false positive rate was achieved with the gradients-based method enhanced with the SimLoss loss function. The energy-based method, which offers faster inference speeds and has a much simpler implementation, achieved an 87.83\% detection rate. This method is becoming a popular baseline for out-of-distribution detection in the computer vision domain, and to the best of our knowledge, we are the first to use it in the network analysis domain.

\subsection{Future work}
\label{sec:future-work}
Even though TLS represents a big portion of encrypted network traffic, we consider the QUIC protocol as the next potential hegemon for the near future. It is becoming the default protocol for major global web services. Unfortunately, the state of public QUIC datasets is even worse than in the TLS datasets area. Therefore, we are going to continue with dataset collection focusing on the QUIC protocol. Our developed toolset is reusable for this task. We plan to study QUIC similarities with TLS. The application of the transfer learning principle on the model trained for TLS to adapt it for QUIC (and vice versa) is a challenging task that would allow for the reusing of machine learning models in the encrypted traffic classification area. 

\section*{Acknowledgements}
This work was supported by the Ministry of the Interior of the Czech Republic, grant No. VJ02010024: ``Flow-Based Encrypted Traffic Analysis,'' and also by the Grant Agency of the Czech Technical University in Prague, grant No. SGS20/210/OHK3/3T/18. Computational resources were supplied by the project ``e-Infrastruktura CZ'' (e-INFRA CZ LM2018140) supported by the Ministry of Education, Youth and Sports of the Czech Republic.

%%
%% The next two lines define the bibliography style to be used, and
%% the bibliography file.
\bibliographystyle{plainnat}
\bibliography{main}

\appendix
    
\section{Detailed classification report}
\label{appendix:classification-report}

\begin{footnotesize}
\begin{verbatim}
                      precision (sc)     recall (sc)   f1-score (sc)
         accuweather     0.99            1.00            0.99       
           adobe-ads     0.94 (0.94)     0.93 (0.93)     0.93 (0.93)
     adobe-analytics     1.00 (1.00)     0.98 (0.98)     0.99 (0.99)
          adobe-auth     0.99 (0.99)     0.99 (0.99)     0.99 (0.99)
         adobe-cloud     0.98 (0.98)     0.99 (0.99)     0.98 (0.99)
 adobe-notifications     0.98 (0.98)     0.99 (1.00)     0.98 (0.99)
        adobe-search     0.99 (1.00)     0.82 (0.99)      0.9 (1.00)
       adobe-updater     0.99 (0.99)     0.98 (0.98)     0.99 (0.99)
  amazon-advertising     0.96            0.96            0.96       
        apple-icloud     0.98 (0.99)     0.97 (0.98)     0.98 (0.99)
        apple-itunes     1.00 (1.00)     0.99 (0.99)     0.99 (1.00)
      apple-location     0.99 (1.00)     0.99 (0.99)     0.99 (0.99)
          apple-ocsp     0.97 (0.99)     0.97 (0.99)     0.97 (0.99)
       apple-pancake     0.99 (1.00)     0.99 (1.00)     0.99 (1.00)
       apple-updates     0.99 (1.00)     0.99 (1.00)     0.99 (1.00)
       apple-weather     0.99 (0.99)     0.99 (0.99)     0.99 (0.99)
            appnexus     0.99            0.99            0.99       
            autodesk     1.00            1.00            1.00       
               avast     0.98            0.98            0.98       
                bing     0.99            0.99            0.99       
   bitdefender-gzone     1.00 (1.00)     1.00 (1.00)     1.00 (1.00)
  bitdefender-nimbus     0.99 (0.99)     0.99 (0.99)     0.99 (0.99)
                chmi     0.99            1.00            0.99       
chrome-remotedesktop     0.92 (1.00)     0.96 (1.00)     0.94 (1.00)
             discord     0.98            0.98            0.98       
             dns-doh     0.92            0.95            0.94       
             dropbox     0.98            0.99            0.99       
            ea-games     0.98            0.99            0.98       
            eset-edf     1.00 (1.00)     1.00 (1.00)     1.00 (1.00)
           eset-edtd     1.00 (1.00)     1.00 (1.00)     1.00 (1.00)
           eset-epns     1.00 (1.00)     1.00 (1.00)     1.00 (1.00)
            eset-esa     1.00 (1.00)     0.99 (0.99)     1.00 (1.00)
             eset-ts     1.00 (1.00)     0.99 (0.99)     1.00 (1.00)
      facebook-graph     0.97 (1.00)     0.97 (1.00)     0.97 (1.00)
      facebook-media     0.92 (1.00)     0.95 (1.00)     0.93 (1.00)
  facebook-messenger     0.98 (1.00)     0.99 (1.00)     0.99 (1.00)
        facebook-web     0.92 (0.99)     0.91 (0.99)     0.92 (0.99)
              github     0.96            0.96            0.96       
               gmail     0.95 (1.00)     0.75 (0.96)     0.84 (0.98)
          google-ads     0.96 (1.00)     0.95 (1.00)     0.95 (1.00)
         google-auth     0.93 (1.00)     0.86 (1.00)     0.89 (1.00)
 google-connectivity     0.99 (1.00)     0.99 (1.00)     0.99 (1.00)
        google-drive     0.89 (1.00)     0.75 (0.99)     0.81 (0.99)
        google-fonts     0.83 (1.00)     0.96 (0.99)     0.89 (0.99)
     google-hangouts     0.97 (1.00)     0.88 (1.00)     0.93 (1.00)
         google-play     0.88 (1.00)     0.91 (0.99)     0.89 (1.00)
 google-safebrowsing     0.95 (1.00)     0.96 (1.00)     0.96 (1.00)
     google-services     0.86 (1.00)     0.79 (0.99)     0.82 (0.99)
    google-translate     0.94 (1.00)     0.81 (1.00)     0.87 (1.00)
 google-userlocation     0.99 (1.00)     0.98 (1.00)     0.98 (1.00)
          google-www     0.97 (1.00)     0.96 (1.00)     0.96 (1.00)
           grammarly     0.98            0.99            0.99       
           instagram     0.97 (1.00)     0.95 (1.00)     0.96 (1.00)
           kaspersky     1.00            0.99            0.99       
          king-games     0.99            0.99            0.99       
              loggly     0.98            0.99            0.99       
  malwarebytes-telem     1.00            1.00            1.00       
              mapscz     0.97 (0.99)     0.92 (0.99)     0.94 (0.99)
          mcafee-ccs     0.99 (0.99)     0.99 (0.99)     0.99 (0.99)
          mcafee-gti     0.99 (0.99)     0.99 (0.99)     0.99 (0.99)
  mcafee-realprotect     0.99 (0.99)     0.99 (0.99)     0.99 (0.99)
   ms-authentication     0.99 (1.00)     0.99 (1.00)     0.99 (1.00)
  microsoft-defender     0.99 (1.00)     1.00 (1.00)     1.00 (1.00)
       ms-diagnostic     1.00 (1.00)     1.00 (1.00)     1.00 (1.00)
     microsoft-notes     0.99 (1.00)     1.00 (1.00)     1.00 (1.00)
  microsoft-onedrive     1.00 (1.00)     0.99 (0.99)     0.99 (0.99)
      microsoft-push     1.00 (1.00)     1.00 (1.00)     1.00 (1.00)
  microsoft-settings     1.00 (1.00)     1.00 (1.00)     1.00 (1.00)
    microsoft-update     1.00 (1.00)     1.00 (1.00)     1.00 (1.00)
   microsoft-weather     0.99 (0.99)     1.00 (1.00)     0.99 (1.00)
                 ndk     0.97            0.94            0.96       
                o2tv     0.99            1.00            1.00       
          office-365     0.97 (0.98)     0.96 (0.98)     0.96 (0.98)
             outlook     0.97 (0.99)     0.98 (0.99)     0.98 (0.99)
            pubmatic     0.98            0.99            0.99       
          riot-games     0.99            0.98            0.99       
         rozhlas-api     0.99            0.99            0.99       
      rubiconproject     0.99            0.99            0.99       
         seznam-auth     0.97 (1.00)     0.97 (1.00)     0.97 (1.00)
        seznam-email     0.99 (1.00)     0.99 (1.00)     0.99 (1.00)
        seznam-media     0.97 (0.99)     0.99 (0.99)     0.98 (0.99)
seznam-notifications     0.96 (1.00)     0.98 (1.00)     0.97 (1.00)
       seznam-search     0.98 (0.99)     0.92 (0.99)     0.95 (0.99)
          seznam-ssp     0.96 (0.99)     0.98 (0.99)     0.97 (0.99)
               skype     0.98 (0.99)     0.97 (0.99)     0.98 (0.99)
            snapchat     0.98            0.98            0.98       
             spotify     0.96            0.98            0.97       
               steam     0.94            0.92            0.93       
               teams     0.97 (0.98)     0.97 (0.98)     0.97 (0.98)
              tiktok     0.98            0.98            0.98       
              twitch     0.97            0.97            0.97       
             twitter     0.98            0.99            0.98       
         unity-games     0.98            0.96            0.97       
            uzis-api     1.00            0.99            1.00       
         vmware-vcsa     0.99            0.99            0.99       
           vse-insis     0.93            0.96            0.94       
            whatsapp     0.95 (0.99)     0.95 (0.99)     0.95 (0.99)
           xbox-live     0.99            0.97            0.98       
             youtube     0.86 (0.99)     0.90 (0.99)     0.88 (0.99)
    
                      precision (sc)     recall (sc)   f1-score (sc)
           macro avg   0.963 (0.982)   0.956 (0.978)   0.959 (0.979)
            accuracy                                   0.974        
 superclass accuracy                                   0.991
\end{verbatim}
\end{footnotesize}

\section{Additional figures}
\begin{figure}[t]
	\centering
	\includegraphics[width=0.75\linewidth]{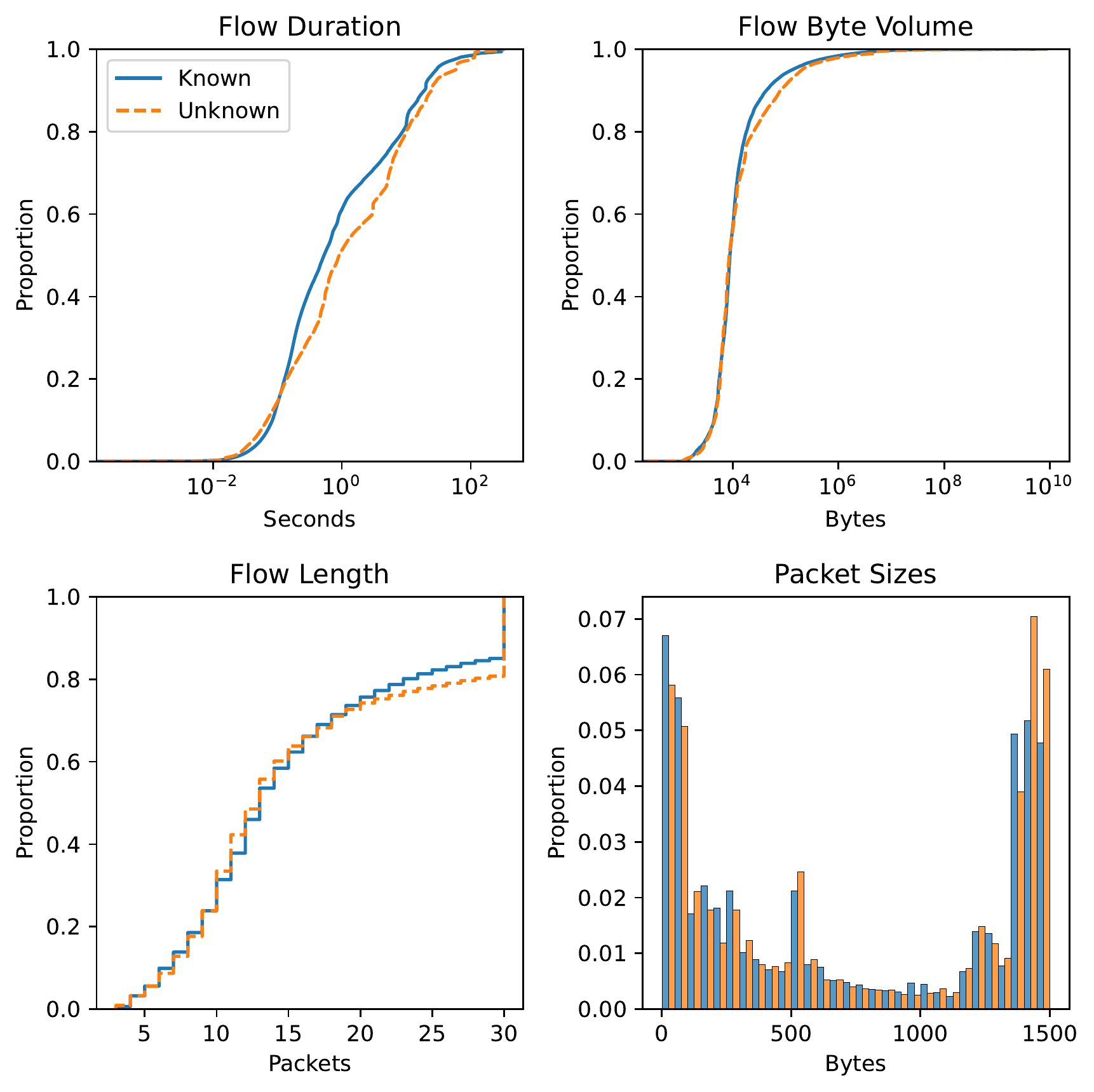}
	\caption{A comparison of traffic characteristics of known versus unknown flows for the top-100 split. The distributions are similar; one noticeable difference is that unknown flows tend to be a bit longer. The median unknown flow duration is 0.9 seconds; for known flows, it is 0.5 seconds. The statistics were computed from 61 million known flows and 6.5 million unknown flows.}
	\label{fig:dataset-stats-known-unknown}
\end{figure}

\begin{figure}
    \centering  
    \begin{subfigure}[b]{1\textwidth}
        \centering
        \includegraphics[width=0.81\linewidth]{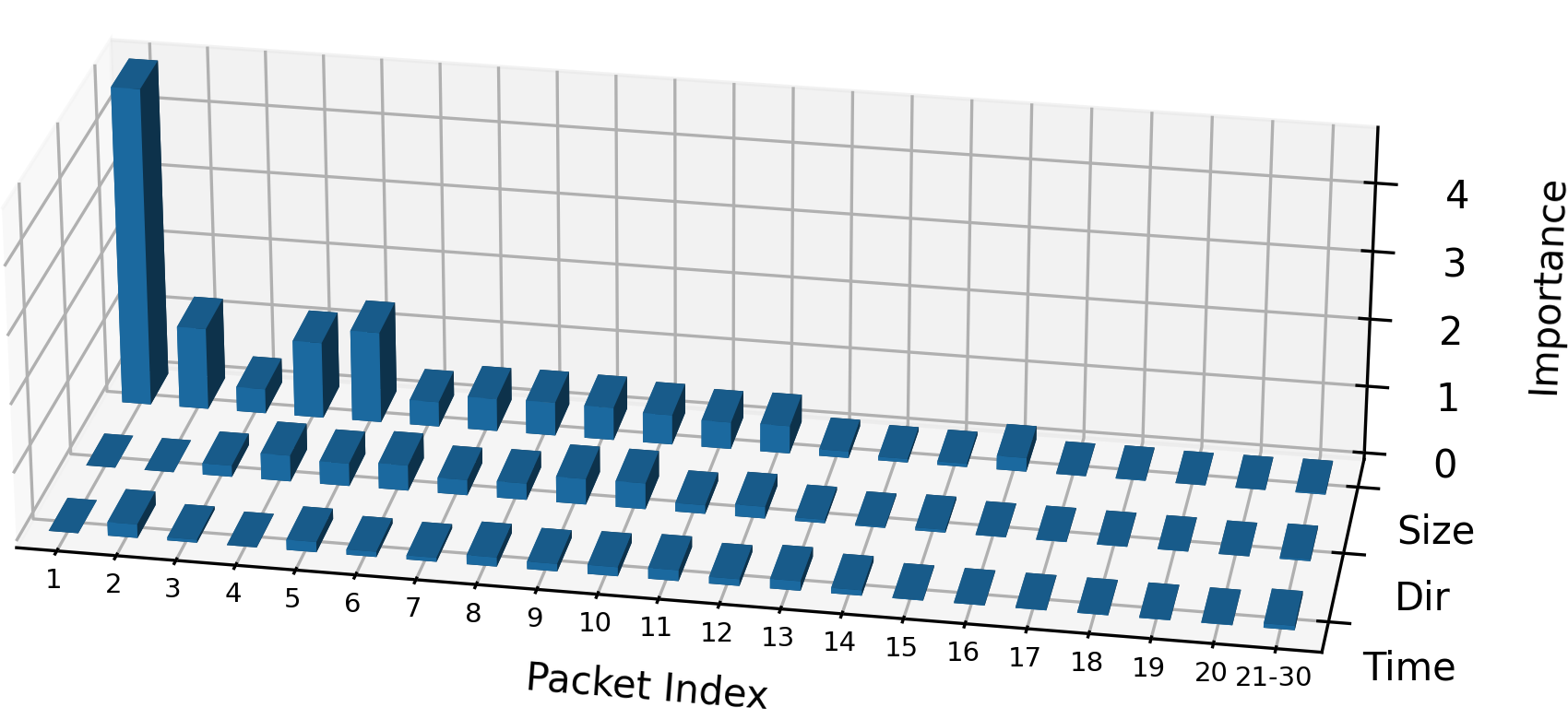}
        \caption{Chrome Remote Desktop API}
    \end{subfigure}
    
    \begin{subfigure}[b]{1\textwidth}
        \centering
        \includegraphics[width=0.81\linewidth]{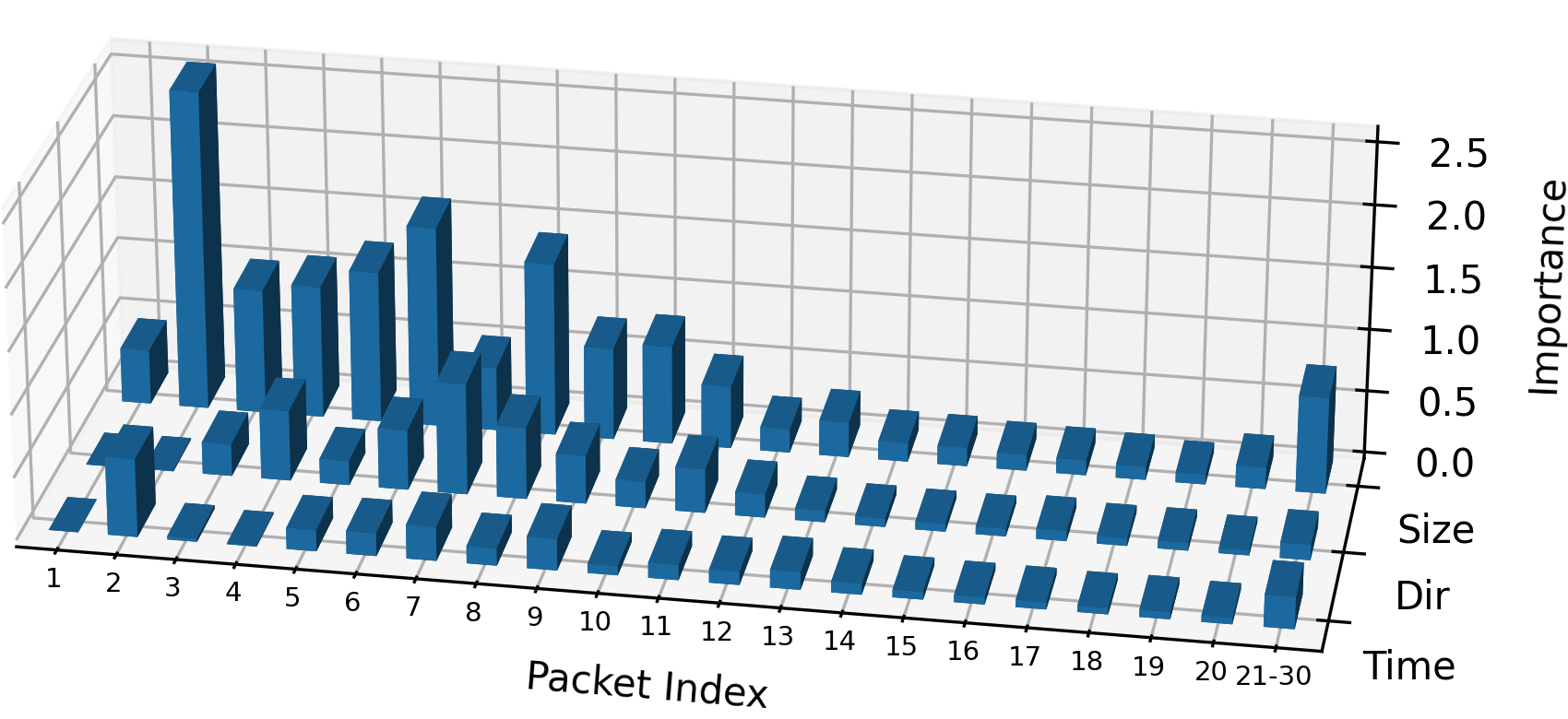}
        \caption{Seznam Notification Service}
    \end{subfigure}
    
    \begin{subfigure}[b]{1\textwidth}
        \centering
        \includegraphics[width=0.81\linewidth]{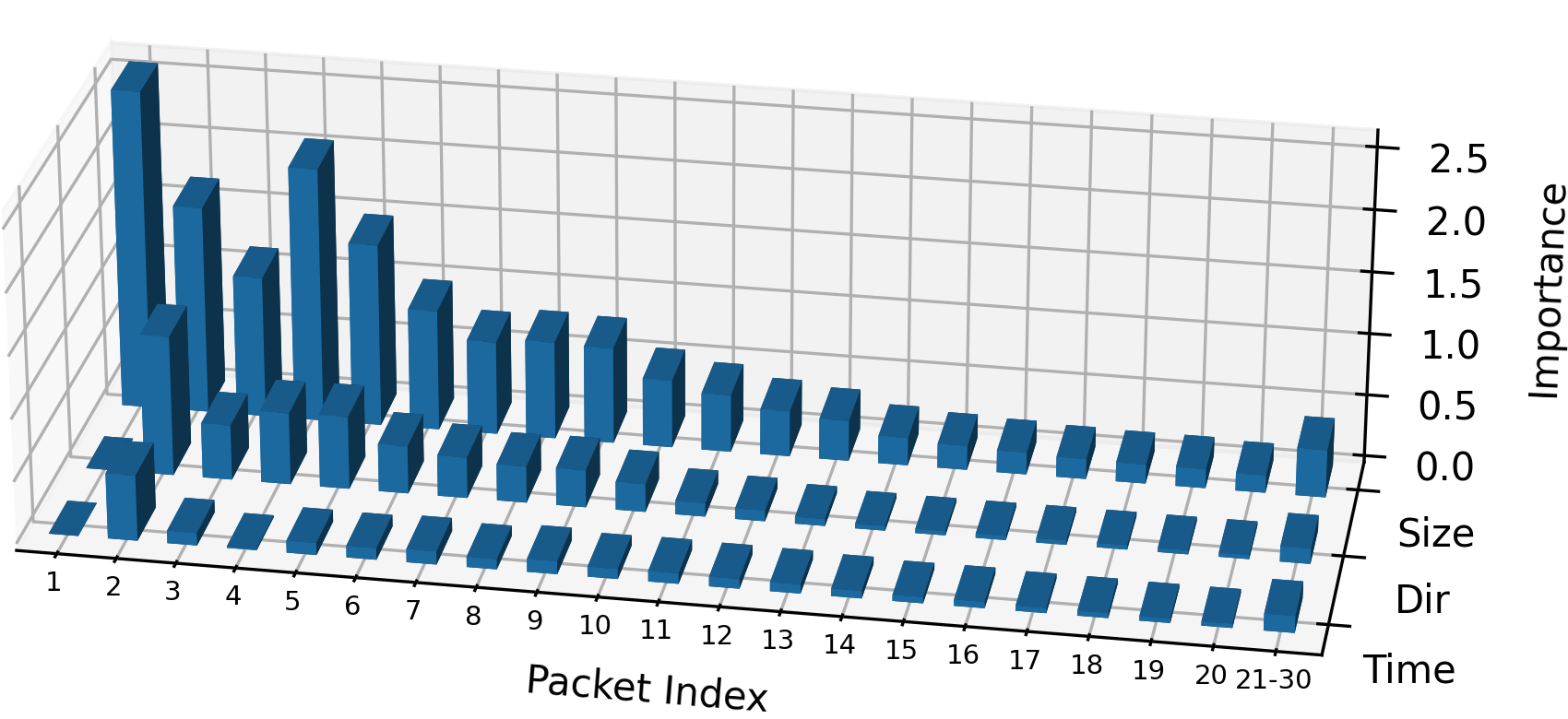}
        \caption{Facebook Messenger}
    \end{subfigure}
\caption{Per-service XAI interpretations. As opposed to Figure~\ref{fig:xai-pstats}, where interpretations are averages across samples of all services, here, we show interpretations for three services to demonstrate that the model relies on different parts of input for different services. (a) For the detection of Chrome Remote Desktop API, the size of the first packet has the utmost importance. (b) The second packet is the most important for the detection of the Seznam Notification Service. Also, packets at positions 21 to 30 still carry valuable information.}
\label{fig:xai-services}
\end{figure}

See Figures~\ref{fig:dataset-stats-known-unknown} and \ref{fig:xai-services}.

\end{document}